\documentclass[journal]{IEEEtran}
\usepackage{cite}

\ifCLASSINFOpdf
  \usepackage[pdftex]{graphicx}
  \graphicspath{{figures/}}
\else
\fi
\usepackage{amsmath}
\usepackage{amssymb}
\DeclareMathOperator*{\argmax}{argmax}
\DeclareMathOperator*{\argmin}{argmin}

\usepackage{algorithm}
\usepackage{algpseudocode}

\algnewcommand\algorithmicforeach{\textbf{for each}}
\algdef{S}[FOR]{ForEach}[1]{\algorithmicforeach\ #1\ \algorithmicdo}
\algdef{SE}[DOWHILE]{Do}{doWhile}{\algorithmicdo}[1]{\algorithmicwhile\ #1}%

\algrenewcommand\algorithmicindent{1.0em}

\ifCLASSOPTIONcompsoc
  \usepackage[caption=false,font=normalsize,labelfont=sf,textfont=sf]{subfig}
\else
  \usepackage[caption=false,font=footnotesize]{subfig}
\fi
\usepackage{booktabs,makecell}
\usepackage{siunitx}

\usepackage{multirow}
\usepackage{hhline}
\usepackage[table,xcdraw]{xcolor}

\usepackage{url}

\makeatletter
\let\NAT@parse\undefined
\makeatother
\usepackage{hyperref}  

\usepackage{array}
\newcolumntype{L}[1]{>{\raggedright\arraybackslash}p{#1}}
\newcolumntype{C}[1]{>{\centering\arraybackslash}p{#1}}
\newcolumntype{R}[1]{>{\raggedleft\arraybackslash}p{#1}}

\usepackage{rotating,siunitx}

\begin{document}
%
\title{Optimized Views Photogrammetry: Precision Analysis and A Large-scale Case Study in Qingdao}
%
%
%

\author{Qingquan~Li,
	Wenshuai~Yu,
	San~Jiang
	\thanks{Q. Li and W. Yu are with the College of Civil and Transportation Engineering, Shenzhen University, Shenzhen 518060, China, and also with the Guangdong Laboratory of Artificial Intelligence and Digital Economy (Shenzhen), Shenzhen 518060, China. E-mail:
	\textit{liqq}@szu.edu.cn, \textit{ywsh}@szu.edu.cn.}
	\thanks{S. Jiang is with the School of Computer Science, China University of Geosciences, Wuhan 430074, China. E-mail: \textit{jiangsan}@cug.edu.cn.}
	\thanks{\textit{Corresponding author: Wenshuai Yu, San Jiang}}}

%
%

\markboth{Journal of \LaTeX\ Class Files}%
{Shell \MakeLowercase{\textit{et al.}}: Bare Demo of IEEEtran.cls for IEEE Journals}
%



\maketitle

\begin{abstract}
UAVs have become one of the widely used remote sensing platforms and played a critical role in the construction of smart cities. However, due to the complex environment in urban scenes, secure and accurate data acquisition brings great challenges to 3D modeling and scene updating. Optimal trajectory planning of UAVs and accurate data collection of onboard cameras are non-trivial issues in urban modeling. This study presents the principle of optimized views photogrammetry and verifies its precision and potential in large-scale 3D modeling. Different from oblique photogrammetry, optimized views photogrammetry uses rough models to generate and optimize UAV trajectories, which is achieved through the consideration of model point reconstructability and view point redundancy. Based on the principle of optimized views photogrammetry, this study first conducts a precision analysis of 3D models by using UAV images of optimized views photogrammetry and then executes a large-scale case study in the urban region of Qingdao city, China, to verify its engineering potential. By using GCPs for image orientation precision analysis and TLS (terrestrial laser scanning) point clouds for model quality analysis, experimental results show that optimized views photogrammetry could construct stable image connection networks and could achieve comparable image orientation accuracy. Benefiting from the accurate image acquisition strategy, the quality of mesh models significantly improves, especially for urban areas with serious occlusions, in which 3 to 5 times of higher accuracy has been achieved. Besides, the case study in Qingdao city verifies that optimized views photogrammetry can be a reliable and powerful solution for the large-scale 3D modeling in complex urban scenes.
\end{abstract}


%
\IEEEpeerreviewmaketitle

\section{Introduction}
\label{sec:1}

\IEEEPARstart{3}{D} reconstruction has become increasingly critical for building smart cities \cite{prandi2014services,yu2021automatic}. Satellite-based and aerial-based photogrammetric imaging systems can provide remote sensing (RS) data for 3D reconstruction of large-scale topographic reliefs and urban buildings \cite{he2022hmsm,zhang2022building}. However, due to their high flight heights, such as hundreds of kilometers on satellites and a few kilometers on aerial planes, and fixed observation viewpoints, these RS systems can only provide data with low spatial resolutions and limited acquisition abilities \cite{wang2021developing}. For reconstructing fine-scale 3D models of complex urban scenes, it has become very urgent to meet the requirements for high timeliness and flexibility in data acquisition as well as high spatial resolution for collected images \cite{kuang2020real,shang2020co}.

In recent years, UAV (unmanned aerial vehicle) platforms have gained extensive attention in varying applications, including but not limited to heritage documentation \cite{murtiyoso2017documentation}, structural monitoring \cite{pan2019three}, and transmission line inspection \cite{jiang2017uav,jiang2019uav}, due to their characteristics of low economic costs, ease to usage, and flexible acquisition abilities. Equipped with consumer-grade digital cameras, UAV-based photogrammetric systems can record images with center-meter-level spatial resolutions. The combination of UAV platforms and oblique photogrammetric techniques can enhance the advantages of flexible and multi-view data acquisition capabilities \cite{jiang2021unmanned}. UAV-based photogrammetry has become an important RS platform in the 3D reconstruction of urban cities \cite{colomina2014unmanned,xiang2019mini}.

\begin{figure*}[!ht]
	\centering
	\subfloat[Model from oblique photogrammetry]{\includegraphics[width=0.4\textwidth]{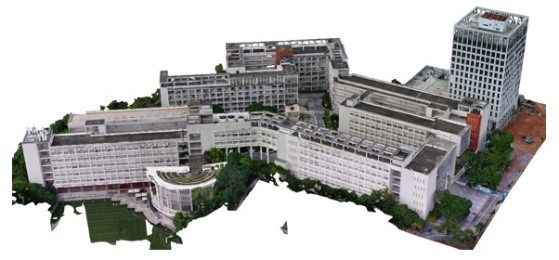}
		\label{fig:figure1-a}}
	\subfloat[Model from optimized views photogrammetry]{\includegraphics[width=0.4\textwidth]{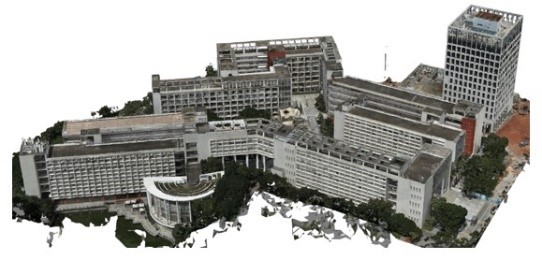}
		\label{fig:figure1-b}} \\
	\subfloat[Comparison of local detail]{\includegraphics[width=0.8\textwidth]{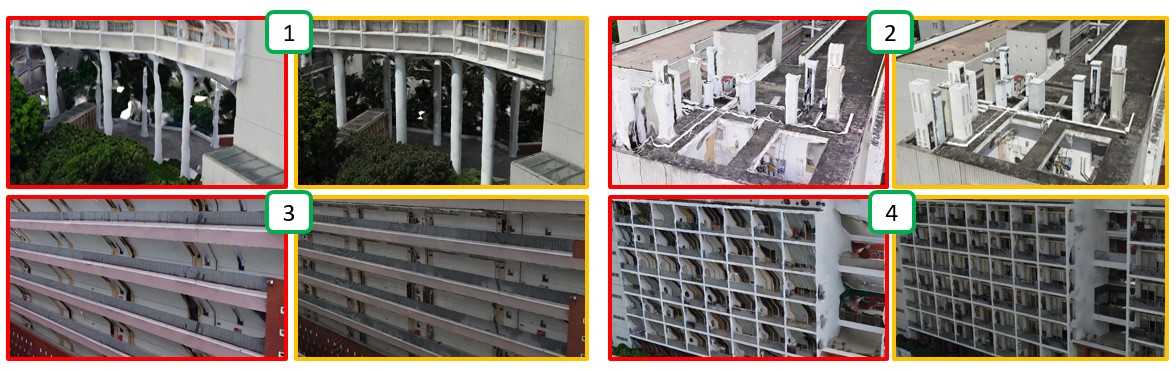}
		\label{fig:figure1-c}}
	\caption{Comparison of mesh models reconstructed by using classical oblique and optimized views photogrammetry.}
	\label{fig:figure1}
\end{figure*}

Generally, the workflow of UAV-based photogrammetry for 3D reconstruction consists of two major steps, i.e., data acquisition and 3D modeling. In the literature, image-based 3D mesh modeling has become mature because of the rapid development of image orientation and multi-view stereo technologies \cite{griwodz2021alicevision}, which can be observed from the widely used commercial and open-source software packages, e.g., Bentley ContextCapture, Pix4Dmapper, ColMap and AliceVision \cite{jiang2020effcient}. On the contrary, data acquisition is not a non-trivial task in complex urban scenes. According to the principle of classical oblique photogrammetry, UAVs are equipped with fixed-orientation cameras and operated at fixed-height trajectories for data acquisitions. Due to the large height variations and serious occlusions of urban buildings, recorded images can only provide very limited observations for 3D modeling \cite{zhang2021continuous}, which leads to the degeneration in both precision and completeness of 3D reconstructed models. Besides, classical photogrammetry uses regular camera exposure positions to record images and cannot consider the geometric characteristics of urban scenes. In other words, both flatten grounds and complex structures are equally scanned, which leads to redundant observations for the former and insufficient observations for the latter \cite{jiang2017Efficient}. Figure \ref{fig:figure1-c} illustrates the local details of reconstructed models from oblique photogrammetry, which are indicated by red boxes. Due to limited observations, fault and deficiency can be seen in building facades and bottoms. Therefore, more accurate trajectory planning methods are required to further improve the model quality in complex urban scenes \cite{zhou2020offsite}.

In the literature, UAV trajectory planning is the technique to design the flying path of UAV platforms and the viewing direction of onboard cameras. To ensure the flight safety and improve the production quality, trajectory planning plays a critical role that needs serious consideration for UAV-based oblique photogrammetry \cite{koch2019automatic,zheng2018multi}. Generally, existing trajectory planning methods can be divided into two major categories, i.e., traditional planning methods and geometry-aware planning methods. In the fields of photogrammetry and remote sensing, traditional planning methods generate trajectories that control UAVs flying above urban scenes at a fixed height in urban areas or by using a ground-adaption mode in mountain regions, and images are collected with vertical or oblique equipped cameras \cite{rupnik2015aerial}. Because of the high altitudes and serious occlusions, images that are captured from these trajectory paths cannot achieve sufficient observations in complex urban scenes \cite{koch2019automatic}.

In contrast to traditional planning methods, geometry-aware planning methods have gained extensive attention in recent years because of the increasing usage of UAVs in the 3D modeling of urban cities \cite{liu2018object}. Through the exploitation of geometric information of ground scenes, geometry-aware planning methods calculate necessary camera viewpoints that are required for accurate 3D reconstruction and generate UAV flying paths with adaptive camera viewing directions. According to the used auxiliary data, existing geometry-aware planning methods can be divided into two-step methods \cite{hepp2018plan3d,koch2019automatic,smith2018aerial} and one-step methods \cite{kuang2020real,zhou2020offsite}. Two-step methods first depend on the traditional photogrammetry acquisition, such as oblique imaging, to collect necessary images of test sites to reconstruct a rough model, which is then used to initialize and refined an optimized trajectory for geometry-aware data acquisition. On the contrary, one-step methods do not depend on any rough models and can directly utilize the prior data of test sites to generate necessary information for trajectory planning. In the work of \cite{zhou2020offsite}, a UAV trajectory planning method, termed optimized views photogrammetry, was designed for the 3D reconstruction of complex urban scenes. Instead of using an extra data acquisition for rough model generation, existing 2D vector maps and RS satellite images have been exploited to generate rough building box models. Combined with dense sampling-based initial viewpoint generation and reconstructability constrained final viewpoint optimization, UAV trajectory planning and accurate data acquisition have been achieved. Figure \ref{fig:figure1-c} illustrates the local details of reconstructed models from optimized views photogrammetry, which are indicated by brown boxes. Compared with conventional trajectory planning methods, reconstructed models from optimized views photogrammetry have higher precision and completeness.

Combined with the high flexibility and multi-view acquisition ability of UAV platforms, optimized views photogrammetry has shown high potential for 3D reconstruction and model updating in complex urban scenes. Thus, this study conducts a comprehensive evaluation of image orientation accuracy and 3D model quality generated from UAV images of optimized views photogrammetry. The primary contributions of this study can be concluded as follows: (1) we present the basic principle of optimized views photogrammetry in the context of urban scene 3D reconstruction and the considerations for engineering applications; (2) we conduct a precision analysis of 3D models that are generated from the UAV images of optimized views photogrammetry by using both GCPs (ground control points) for image orientation accuracy analysis and TLS (terrestrial laser scanning) point clouds for mesh model quality analysis; (3) we conduct a large-scale case study in the urban region of Qingdao city, China, to verify the performance of optimized views photogrammetry in engineering applications.

This paper is organized as follows. Section \ref{sec:2} gives the mathematical basics of optimized views photogrammetry. Section \ref{sec:3} presents the workflow and detailed procedure of optimized views photogrammetry. By using oblique and optimized views photogrammetry, two test sites are selected for comprehensive performance evaluation in Section \ref{sec:4}. Finally, Section \ref{sec:5} presents the conclusion of this study.

\section{Mathematic basic of optimized views photogrammetry}
\label{sec:2}

The main purpose of optimized views photogrammetry is to define a set of view points, from which model points obtain enough observation with minimal view point redundancy. From the aspect of multi-view stereo reconstruction, optimized views photogrammetry aims to minimize the redundancy of selected view points and maximize the reconstructability of reconstructed model points. In contrast to classical oblique photogrammetry that focuses on the stability of image connection networks, optimized views photogrammetry attempts to achieve the 3D reconstruction of urban scenes with high precision and completeness.

\begin{figure}[t!]
	\centering
	\subfloat{\includegraphics[width=0.4\textwidth]{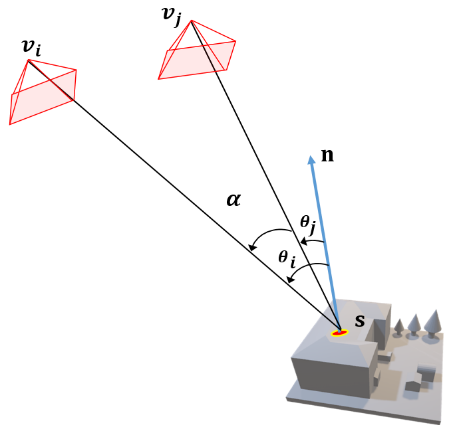}
		\label{fig:figure2}}
	\caption{The illustration of reconstructability of model points under selected view points.}
	\label{fig:figure2}
\end{figure}

According to the notations presented in \cite{smith2018aerial}, reconstructability $q(s,v_i,v_j)$ defines the reconstruction quality of model point $s$ under two selected view points $(v_i,v_j)$, as illustrated in Figure \ref{fig:figure2}. Reconstructability $q(s,v_i,v_j)$ is calculated using Equation \ref{eq:1}

\begin{equation}
	q(s,v_i,v_j)=w_1(\alpha)w_2(d_m)w_3(\alpha)cos(\theta_m)
	\label{eq:1}
\end{equation}
where $\alpha$ is the intersection angle of view points $(v_i,v_j)$ when seeing model point $s$; $d_m=max(\left \| sv_i \right \|,\left \| sv_j \right \|)$ indicates the maximum distance from view points $(v_i,v_j)$ to model point $s$; $\theta _m=max(\theta _i,\theta _j)$ defines the maximal intersection angle between vectors from model point $s$ to view points $(v_i,v_j)$ and the normal vector of model point $s$; $w_1$, $w_2$ and $w_3$ are the weight items. Thus, reconstructability $q(s,v_i,v_j)$ quantifies the relationship between view points and model points under three constraints, i.e., the intersection angle of view points, the imaging distance from view points to model point and deviation from the normal vector of model point. To obtain high reconstructability $q(s,v_i,v_j)$, view points $(v_i,v_j)$ with medium intersection angle, near imaging distance and direction to the model point and its normal vector, respectively, are preferred.

Based on the definition of reconstructability $q(s,v_i,v_j)$ of model point $s$ under two view point $(v_i,v_j)$, the reconstructability of model point $s$ under the view point set $U$ is defined as the sum of $q(s,v_i,v_j)$ between all possible view point pairs, which can be calculated by using Equation \ref{eq:2}
\begin{equation}
	h(s,U)=\sum_{ i=1,...,|U|, j=i+1,...,|U|} \delta (s,v_i) \delta (s,v_j) q(s,v_i,v_j)
	\label{eq:2}
\end{equation}
where $\delta (s,v_i)$ and $\delta (s,v_j)$ define the visibility of model point $s$ under view points $(v_i,v_j)$. It is set as one if model point $s$ is visible in the corresponding view point; otherwise, it is set as zero. Similarly, the redundancy $r(v)$ of view point $v$ under the view point set $U$ can be calculated by using Equation \ref{eq:3}
\begin{equation}
	r(v, U)=min \{ h(s,U) | s \in S, \delta (s,v) \}
	\label{eq:3}
\end{equation}
where $S$ defines the set of model points that can be observed by view point $v$. Thus, the redundancy $r(v)$ is defined as the minimum reconstructability $h(s,U)$ of all observed model points $S$ under the view point set $U$.

\section{Workflow of optimized views photogrammetry}
\label{sec:3}

\begin{figure*}[t!]
	\centering
	\includegraphics[width=0.8\textwidth]{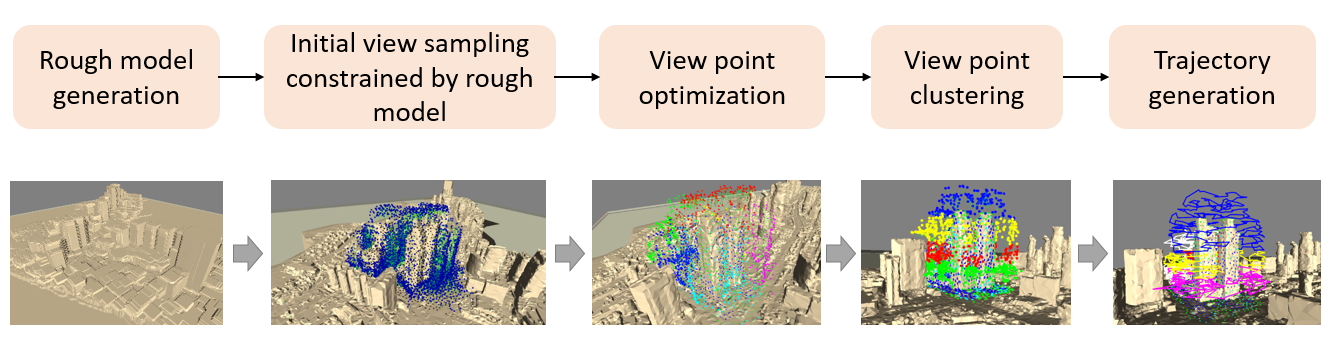}
	\caption{The workflow of optimized views photogrammetry.}
	\label{fig:figure3}
\end{figure*}

The workflow of optimized views photogrammetry consists of five steps, as shown in Figure \ref{fig:figure3}, including rough model generation, initial view point sampling constrained by rough model, view point optimization based on reconstructability analysis, view point clustering and trajectory generation \cite{zhou2020offsite}. The input is the rough model of the test site, which includes the basic geometric priors of the test site. Based on the rough model, initial view points are then generated by the dense sampling on the rough model. Since the generation of initial view points does not consider the efficiency of data acquisition, serious redundancies exist and would cause huge economic costs for data acquisition. Thus, initial view points are further optimized by selecting a small enough subset of view points, which minimizes the redundancy of view points and maximizes the reconstructability of model points. Since one data acquisition campaign of UAVs cannot cover the whole trajectory path, refined view points are divided into different clusters, and the view points in each cluster are used to generate an individual trajectory path, which covers one part of the entire test site. The details of each step are presented in the following subsections.

\subsection{Rough model from prior data or oblique photogrammetry}
\label{sec:3.1}

The rough model includes the geometric information of test sites, which acts as the basis for optimized views photogrammetry. In general, the rough model can be obtained by using three strategies. For the first one, existing 2D geo-spatial data can be directly utilized, such as 2D vector maps with height attributes to build 2.5D box models. For the second one, design data in the field of BIM (building information model), such as CAD (computer-aided design) models, can be used to generate rough models after the processing of geo-referencing and similarity transformation. Although these two methods do not depend on other data acquisition campaigns, the reality of the generated rough models can not be ensured, which may occur frequently in build-up regions. To obtain reliable rough models, the third strategy uses classical oblique photogrammetry to record necessary images for the reconstruction of rough models.

\begin{figure}[t!]
	\centering
	\subfloat[entity object model]{
		\includegraphics[width=0.23\textwidth]{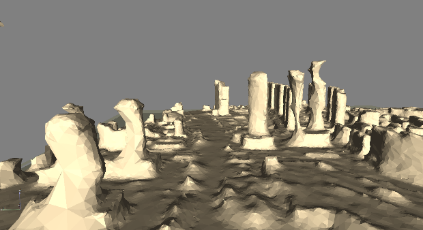}
		\label{fig:figure4-a}
	}
	\subfloat[hierarchical 2.5D model]{
		\includegraphics[width=0.23\textwidth]{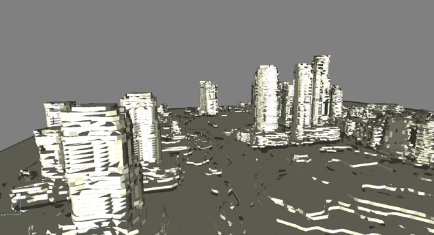}
		\label{fig:figure4-b}
	}
	\caption{Entity object model and hierarchical 2.5D model.}
	\label{fig:figure4}
\end{figure}

The urban scene usually covers a large-scale region and includes a variety of ground objects. Besides, high-frequency update changes occur in urban cities. To obtain the reality, the third strategy has been adopted in this study for rough model generation. By using images collected from an extra acquisition campaign, optimized view photogrammetry generates two kinds of rough models through 3D reconstruction, which are termed as entity object model and hierarchical 2.5D model. Figure 4 shows two kinds of rough models of urban scenes, in which Figure \ref{fig:figure4-a} and Figure \ref{fig:figure4-b} show the entity object model and hierarchical 2.5D model. The entity object model is directly reconstructed from the collected images through 3D reconstruction. On the contrary, the hierarchical 2.5D model is generated hierarchically. Dense point clouds are first divided into different parts based on their height range, and a partial model is reconstructed for each divided point cloud. The final 2.5D model is the concatenation of all partial models.

The practical application indicates that although the structure of entity object models is not complete compared with hierarchical 2.5D models, the entity object model can better reflect the details of ground objects, especially for some special-shaped buildings. On the contrary, the hierarchical 2.5D models are difficult to represent the special structures, such as hollows on buildings. Thus, optimized views photogrammetry uses the entity object model to represent rough models for UAV trajectory planning.

\begin{figure}[t!]
	\centering
	\subfloat[Rough model]{
		\includegraphics[width=0.23\textwidth]{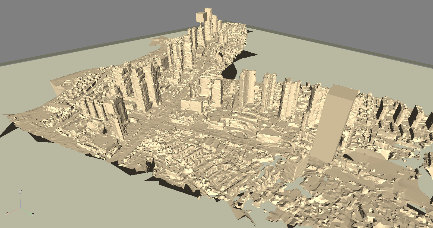}
		\label{fig:figure5-a}
	}
	\subfloat[Safe-flying zone]{
		\includegraphics[width=0.23\textwidth]{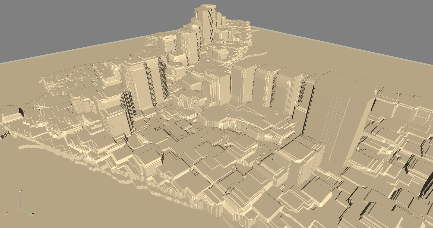}
		\label{fig:figure5-b}
	}
	\caption{Safe-flying zone generation from the rough model.}
	\label{fig:figure5}
\end{figure}

\subsection{Initial view point sampling constrained by rough models}
\label{sec:3.2}

The rough model includes the basic geometric information of test sites, such as buildings, and vegetation in urban scenes, with which the planned trajectory of UAVs could not collide. Due to the missing of some fine-scale objects and the loss of navigation signals in dense building regions, as shown in Figure \ref{fig:figure5-a}, the rough model can not define the safe-flying zone of UAVs. In the optimized view photogrammetry, the dilation of the initial rough model is conducted to define the safe-flying zone, which considers extra uncertainties, such as the incompleteness of the rough models, and the degeneration of navigation precision. Figure \ref{fig:figure5-b} shows the safe-flying zone generated of the rough model, as presented in Figure \ref{fig:figure5-a}.

\begin{figure}[t!]
	\centering
	\includegraphics[width=0.4\textwidth]{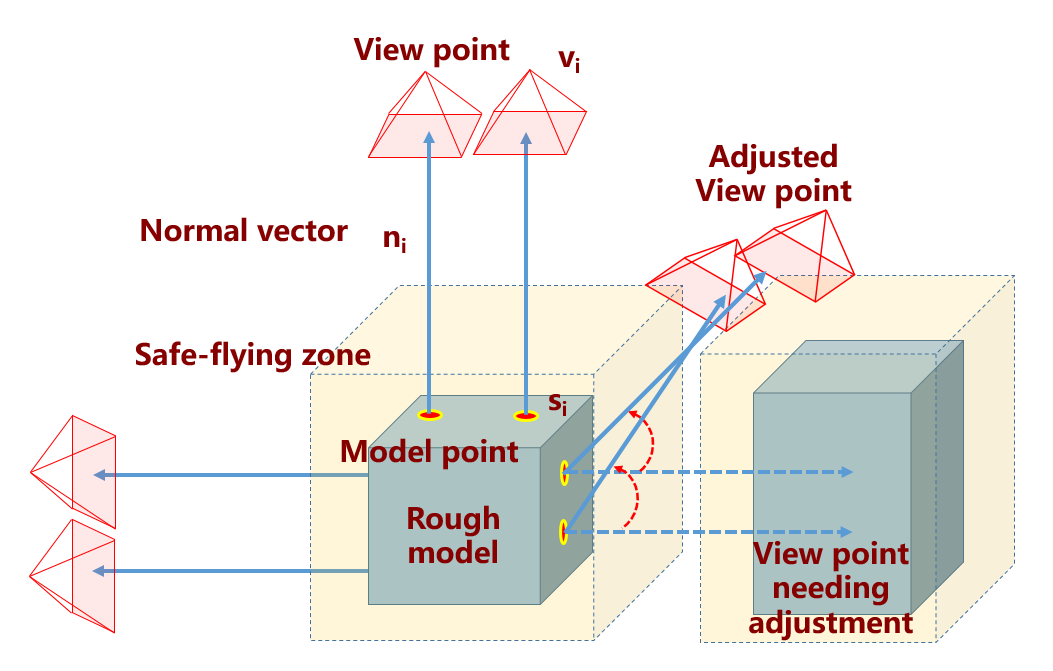}
	\caption{Initial view point generation constrained by rough models.}
	\label{fig:figure6}
\end{figure}

Initial view points can be then sampled by using the constraint of the safe-flying zone. The principle of initial view point sampling is illustrated in Figure \ref{fig:figure6}. First, dense model points $S=\{s_i,\boldsymbol{n_i} \}$ are sampled from the surface of the rough model by using the Poisson disk sampling algorithm \cite{corsini2012efficient}, in which $\boldsymbol{n_i}$ is the normal vector of the model point $s_i$. Compared with other sampling methods, the Poisson disk sampling algorithm focuses on the important regions that reflect the distinctive characteristics the of the rough models. After the sampling of model point $s_i$, its corresponding view point $v_i$ is determined by moving model point $s_i$ along the direction of  normal vector $\boldsymbol{n_i}$ with the distance of $d_{GSD}$, and the imaging direction of view point $v_i$ is opposite to the normal vector $\boldsymbol{n_i}$, as shown in Figure \ref{fig:figure6}. Based on the above-mentioned operation, initial view points $V=\{v_i,\boldsymbol{o_i} \}$ can be obtained, which meets the requirements as represented by Equations \ref{eq:4} and \ref{eq:5}
\begin{equation}
	v_i=s_i+d_{GSD}*\boldsymbol{n_i}
	\label{eq:4}
\end{equation}
\begin{equation}
	\boldsymbol{o_i} = -\boldsymbol{n_i}
	\label{eq:5}
\end{equation}

During Poisson disk sampling, the dimension $d_{GSD}$ of the sampling disk determines the overlap degree $r_{overlap}$ of images that are recorded at view points $V$. The mathematic relationship between the sampling disk and the overlap degree is formulated by Equations \ref{eq:6} and \ref{eq:7}, in which $\theta$ is the field of view (FOV) of the used cameras. Thus, the sampling density of initial view points is controlled by the desired image overlap degree.
\begin{equation}
	D_{disk}=d_{prj}*(1.0-r_{overlap})
	\label{eq:6}
\end{equation}
\begin{equation}
	d_{prj}=2*d_{GSD}*tan(\theta/2)
	\label{eq:7}
\end{equation}

\begin{figure}[t!]
	\centering
	\subfloat[model sampling points]{
		\includegraphics[height=0.15\textwidth]{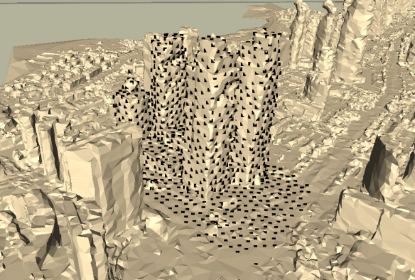}
		\label{fig:figure7-a}
	}
	\subfloat[initial view points]{
		\includegraphics[height=0.15\textwidth]{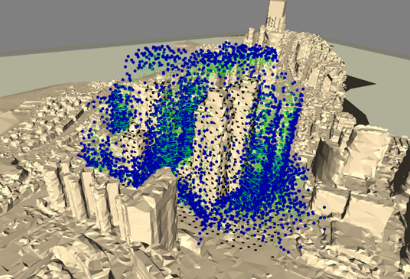}
		\label{fig:figure7-b}
	}
	\caption{The illustration of model sampling points and initial view points.}
	\label{fig:figure7}
\end{figure}

Because of the occlusion of neighboring objects, a subset of initial view points maybe intersect with other geometry objects or within the safe-flying zone, as shown in Figure \ref{fig:figure6}. For this situation, the initial view points are rotated until going out of the safe-flying zone. Figure \ref{fig:figure7} shows an example of model sampling points and initial view points, in which Figure \ref{fig:figure7-a} and Figure \ref{fig:figure7-b} are the model sampling points and initial view points, respectively. We can see that the model sampling points rendered by black rectangles are sparsely located on the rough model, whose sampling density is correlated to the geometric structure of the rough model. Initial view points are rendered by blue circles with their imaging directions indicated by green lines, as presented in Figure \ref{fig:figure7-b}. For this model, a total number of 5,709 view points are generated from the rough model.

\subsection{View point optimization considering redundancy and constructability}
\label{sec:3.3}

Extremely high redundancy exists in the initial view points as they are merely generated based on the dense sampling from the rough model, as illustrated in Figure \ref{fig:figure7-b}, which can not be directly used for trajectory planning in optimized views photogrammetry. Based on the mathematical basic of optimized views photogrammetry as presented in Section \ref{sec:2}, view point optimization is then executed to reduce view point redundancy while maximizing the reconstructability of model points simultaneously.

According to the definition of reconstructability $h(s,U)$ of model point $s$ and the redundancy $r(v)$ of view point $v$ under the initial view points $U$, as presented by Equations \ref{eq:2} and \ref{eq:3}, the purpose of view point optimization is to select a subset of view points $V$ from initial view points $U$, which simultaneously minimizes view point redundancy $R(V)=\sum_{v \in V} r(v)$ and maximizes model point reconstructability $H(S,V)=\sum_{s \in S} h(s,V)$. Thus, the objective function for view point optimization is represented by Equation \ref{eq:8}, in which the threshold $t_h$ determines the minimum reconstructability of model points.
\begin{equation}
\begin{split}
		V^*=\argmax_{V \in U,|V|=|W|} H(S, \argmin_{W \in U}{R(W)}) \\
		s.t. \; h(s,V) > t_h, h(s,W) > t_h, \forall s \in S
	\label{eq:8}
\end{split}
\end{equation}

\begin{figure}[t!]
	\centering
	\includegraphics[width=0.4\textwidth]{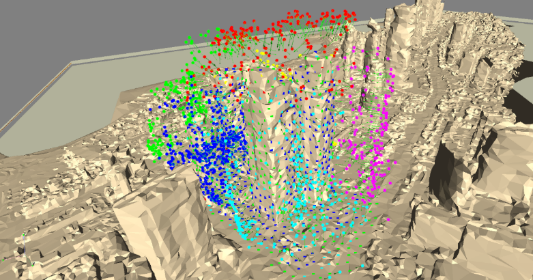}
	\caption{The illustration of optimized view points.}
	\label{fig:figure8}
\end{figure}

The solution to the optimization problem can be achieved through a two-step algorithm: (1) the first step is to minimize the view point redundancy, which is achieved by iteratively removing the view point $v$ with the maximum redundancy $r(v)$ and updating the reconstructability of all related model points. If the reconstructability of at least one model points is less than the threshold $t_h$, then rollback the view point deletion operation and step to the next redundant view point. After the deletion operation, a reduced view point set $V_r$ can be generated; (2) the second step is to maximize the model point reconstructability. For each view point $v_i$ in $V_r$, find its neighboring and similar view point set $\Omega (v_i)$ from initial view points $V$. Iterating over each view point in $\Omega (v_i)$ and finding the view point $v_i^*$ that can maximize the model point reconstructability $H(S,U)$. If the view point $v_i^*$ exist, replacing $v_i$ with $v_i^*$. After the substitution operation, optimized view points $V^*=\{ v_i^*, \boldsymbol{o_i^*}\}$ can be obtained. Figure \ref{fig:figure8} shows the optimized view points from Figure \ref{fig:figure7-b}.

\subsection{View point clustering and UAV trajectory generation}
\label{sec:3.4}

After view point optimization, flying positions of UAV platforms and imaging directions of onboard cameras can be determined and utilized for data acquisition campaigns. In practical application, discrete view points should be stringed orderly to create the trajectory and determine the flying path of UAV platforms. Taking the view points as the waypoints, the trajectory generation of UAVs can be cast as a static path planning problem. It can be formulated as a TSP (traveling salesman problem) optimization problem and could be solved by minimizing the total costs of visiting all view points. During UAV data acquisition, the edge cost between two view points $v_i$ and $v_j$ can be formulated by using their edge length $l(v_i,v_j)$ and imaging angle $\theta (v_i,v_j)$, as formulated in Equation \ref{eq:9}
\begin{equation}
	e(v_i,v_j)=l(v_i,v_j) \exp^{\frac{\theta (v_i,v_j)}{l(v_i,v_j)}}
	\label{eq:9}
\end{equation}

\begin{figure}[t!]
	\centering
	\subfloat[trajectory from view direction clustering]{
		\includegraphics[height=0.15\textwidth]{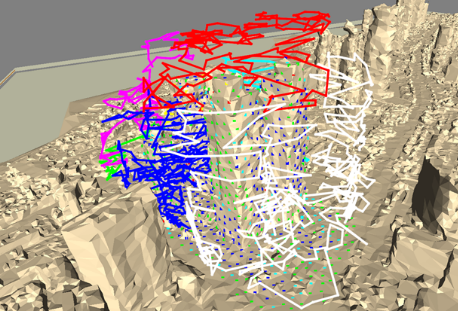}
		\label{fig:figure9-a}
	}
	\subfloat[trajectory from height hierarchical clustering]{
		\includegraphics[height=0.15\textwidth]{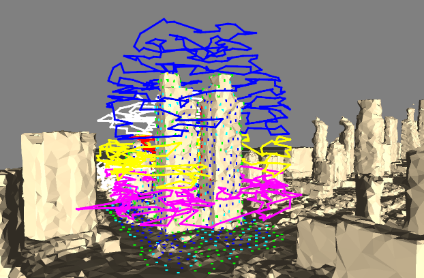}
		\label{fig:figure9-b}
	}
	\caption{The illustration of UAV trajectory generation.}
	\label{fig:figure9}
\end{figure}

Due to a large number of view points and the consideration of building occlusions, the solving of the TSP problem is very difficult and low efficient. In optimized views photogrammetry, this optimization problem is solved by using the genetic algorithm, which could provide a sub-optimal solution with high efficiency. In addition, optimized view points are first clustered into different groups based on their spatial distance, view direction, and height distribution. Figure \ref{fig:figure9} shows the cluster results by using the view direction and height distribution, respectively. In practical applications of optimized views photogrammetry, trajectory path is generated by using the clusters of view points due to two main reasons. On the one hand, the inner structure of each cluster can be consistent; on the other hand, the complexity of optimization solving can be decreased.

\section{Experiment and results}
\label{sec:4}

In the experiments, two test sites are selected for the precision analysis and large-scale engineering application of optimized views photogrammetry. For the precision analysis, both ground control points (GCPs) and terrestrial laser scanning (TLS) point clouds have been surveyed in the first test site, and the performance of optimized views photogrammetry has been evaluated in terms of image ground coverage, tie-point distribution, and relative and absolute bundle adjustment (BA). For the real engineer application, a large-scale case study has been conducted in Qingdao city, China, in which the 3D models of the building-up areas have been scanned and reconstructed based on optimized views photogrammetry.

\subsection{Test sites and datasets}
\label{sec:4.1}

Two test sites have been selected for the evaluation and verification of optimized views photogrammetry. For UAV image acquisition, detailed information for flight configuration is listed in Table \ref{tab:table1}. Noticeably, in these two test sites, both conventional Penta-view oblique and proposed optimized views photogrammetry have been conducted for data acquisition. For the first test site, oblique images are collected for performance evaluation; on the contrary, the oblique images of the second test site are used for rough model generation.

\begin{table*}[!ht]
	\centering
	\caption{Detailed information for flight configuration of the two test sites.}
	\label{tab:table1}
	\makebox[\linewidth]{
		\begin{tabular}{lllll}
			\toprule
			\multirow{2}{*}{Item Name} & \multicolumn{2}{l}{Test Site 1} & \multicolumn{2}{l}{Test Site 2} \\
			\cline{2-5}
			& Oblique & Optimized & Oblique & Optimized \\
			\midrule
			UAV type & multi-rotor & multi-rotor & multi-rotor & multi-rotor \\
			Flight height (m) & 100 & 80 & 340 & 70 \\
			Forward/side overlap (\%) & 85 & 85 & 85 & 85 \\
			Camera mode & PSDK 102S & DJI Zenmuse P1 & \begin{tabular}[c]{@{}l@{}}PhaseOne\\ iXM-RS150F\end{tabular} & DJI FC6310R \\
			Number of cameras & 5 & 1 & 5 & 1 \\
			Focal length (mm) & 35 & 35 & 40/70 & 24 \\
			Camera mount angle (°) & \begin{tabular}[c]{@{}l@{}}Nadir: 0\\ Oblique: 45\end{tabular} & - & \begin{tabular}[c]{@{}l@{}}Nadir: 0\\ Oblique: 45\end{tabular} & - \\
			Number of images & 3620 & 4030 & 2433 & 78640 \\
			Image size (pixel×pixel) & 6000×4000 & 8192×5460 & 14204x10652 & 5472×3648 \\
			GSD (cm) & 1.6 & 1.0 & 2.8 & 2.6 \\
			\bottomrule
		\end{tabular}
	}
\end{table*}

\begin{figure}[t!]
	\centering
	\subfloat[GCP distribution]{
		\includegraphics[height=0.15\textwidth]{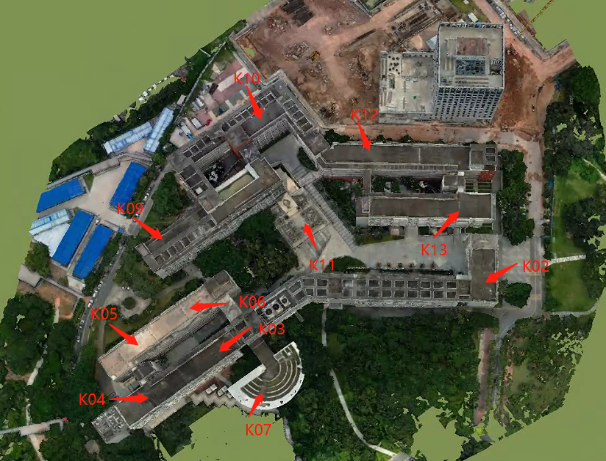}
		\label{fig:figure10-a}
	}
	\subfloat[GCP in building top and facade]{
		\includegraphics[height=0.15\textwidth]{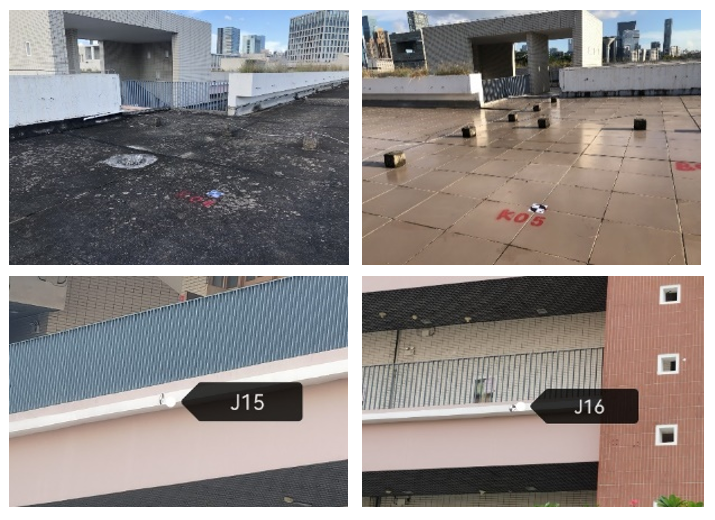}
		\label{fig:figure10-b}
	}
	\caption{GCP distribution and samples in the first test site.}
	\label{fig:figure10}
\end{figure}

The first test site locates in the YueHai distinct of ShenZhen University, as shown in Figure \ref{fig:figure10-a}. There exists a complex building with a corridor-like structure. Since some parts are very close to each other, occlusions can be observed between building parts. The average building height is about 35 m, and a higher office building with a height of 55 m exists near the top-right region. In this test site, other regions are mainly covered by dense vegetation. For data acquisition, a DJI M300 multi-rotor UAV has been adopted in this test site for both oblique and optimized views photogrammetry. The details are described as follows:

\begin{itemize}
	\item For oblique photogrammetry, the multi-rotor UAV is equipped with one PSDK 102S Penta-view imaging system, which is mounted with 0° and 45° respectively for the nadir and oblique cameras. Besides, the focal length of cameras is 35 mm. Under the fixed flight height of 100 m, a total number of 3620 images have been collected with the dimensions of 6000 by 4000 pixels. The mean GSD (Ground Sampling Distance) is approximately 1.6 cm, and the overlap degree of images is about 85\%.
	\item For optimized views photogrammetry, the multi-rotor UAV is equipped with one DJI Zenmuse P1 camera with the dimensions of 8192 by 5460 pixels. In contrast to the fixed angles in oblique photogrammetry, the imaging angles of optimized views are adjusted adaptively to accommodate the geometric structure of ground objects. Under the view distance of 80 m, 4030 images are recorded at this test site.
\end{itemize}

In the first test site, both ground control points and terrestrial laser scanning point clouds have been collected as ground-truth data for geo-reference accuracy and model precision analysis. In this test site, a total number of 28 GCPs have been surveyed by using the UFO-U5 RTK-GNSS (Real-Time Kinematic GNSS) made by the UniStrong cooperation, whose nominal accuracies are 0.8 cm and 1.5 cm in the horizontal and vertical directions, respectively. As presented in Figure \ref{fig:figure10}, the selected GCPs are evenly distributed in the test site, including plain ground, building top and facade.

\begin{figure}[t!]
	\centering
	\includegraphics[width=0.45\textwidth]{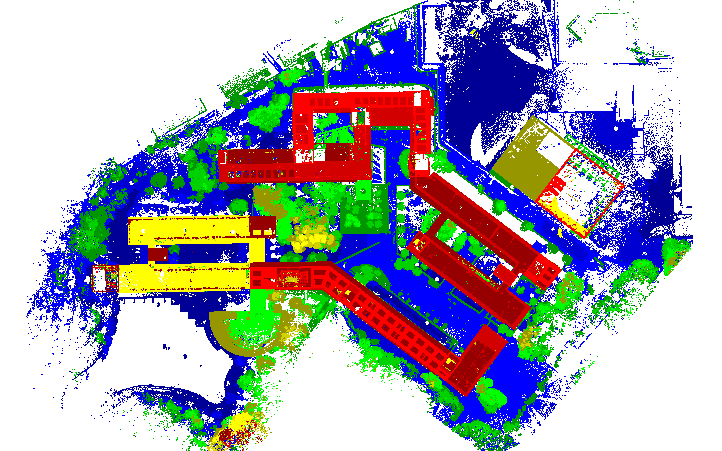}
	\caption{Terrestrial laser scanning point clouds of the first test site.}
	\label{fig:figure11}
\end{figure}

To evaluate the model precision, terrestrial laser scanning point clouds have also been surveyed at this test site. Considering that one survey station cannot cover the whole test site, several survey stations have been set by using the corresponding GCPs as controls, and the point clouds of the whole test site are obtained through accurate registration, as shown in Figure \ref{fig:figure11}. The used instrument is a Trimble X7 laser scanner, whose finding range is 80 m with an accuracy better than 2.4 mm within the scanning distance of 20 m.

\begin{figure}[t!]
	\centering
	\includegraphics[width=0.4\textwidth]{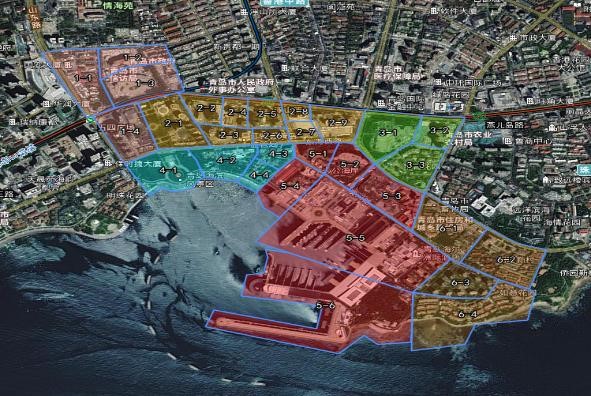}
	\caption{The ground coverage and task division in the second test site.}
	\label{fig:figure12}
\end{figure}

The second test site locates in Qingdao city, China, which is used for the real engineer application of optimized views photogrammetry. The coverage of this test site is about 2.1 square kilometers with small topography relief. Since its a central business district, there are many buildings with large height variations. The highest building is about 223 m. The ground coverage of this test site is presented in Figure \ref{fig:figure12}. In this test site, both oblique and optimized views photogrammetry have been conducted. The former is used to build rough models; the latter is used to collect images for urban 3D model reconstruction.

\begin{itemize}
	\item For oblique photogrammetry, the DJI M300 multi-rotor UAV has been used, which is equipped with a PhaseOne iXM-RS150F Penta-view imaging system. The nadir and oblique cameras are set as 0° and 45°, respectively. Under the fixed height of 340 m, a total number of 2433 images with the dimensions of 14204 by 10652 pixels has been collected for the whole test site. The mean GSD is about 2.8 cm.
	\item For optimized views photogrammetry, six DJI Phantom RTK UAVs have been used to achieve parallel data acquisition. Considering the characteristics of trajectories from optimized views photogrammetry and the endurance of UAVs, the whole test site is first divided into six zones with labels from one to six, and each zone is then further divided into some sub-zones based on the areas of buildings, as shown in Figure \ref{fig:figure12}. The area of the largest zones, i.e., 5-6, is about 0.37 square kilometers, and the areas of the smallest zones, i.e., 2-8, are near 0.02 square kilometers. In this test site, a total number of 78640 images have been collected.
\end{itemize}

\subsection{Analysis of trajectory planning and model point reconstructability}
\label{sec:4.2}

Model point reconstructability $h(s,U)$ is the most important mathematic basic for the design and implementation of optimized views photogrammetry, which indicates how well the model point $s$ would be reconstructed under the selected view point set $U$. It can be considered a useful indicator to evaluate the quality of the subsequent 3D models under a specified trajectory planning. In this subsection, we would analyze the trajectory planning between oblique and optimized views photogrammetry and further evaluate model point reconstructability under their corresponding trajectory configurations.

\begin{figure*}[t!]
	\centering
	\subfloat[oblique photogrammetry]{
		\includegraphics[width=0.4\textwidth]{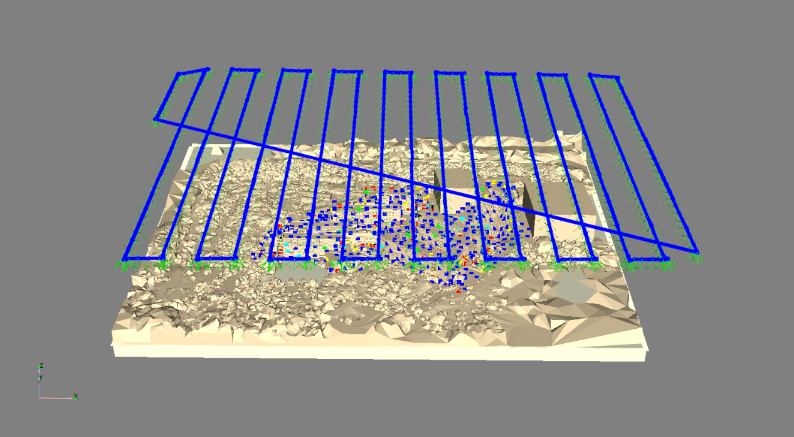}
		\label{fig:figure13-a}
	}
	\subfloat[optimized views photogrammetry]{
		\includegraphics[width=0.4\textwidth]{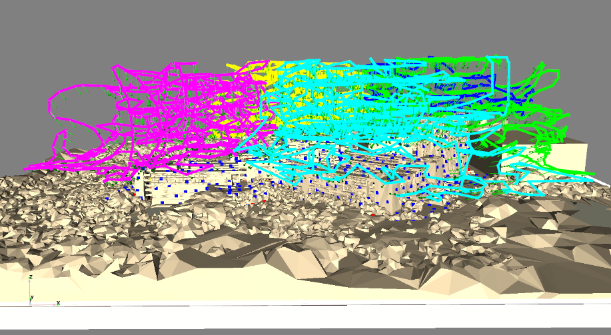}
		\label{fig:figure13-b}
	}
	\caption{The comparison of trajectory planning between oblique and optimized views photogrammetry in the first test site.}
	\label{fig:figure13}
\end{figure*}

\begin{figure*}[t!]
	\centering
	\subfloat[oblique photogrammetry]{
		\includegraphics[width=0.4\textwidth]{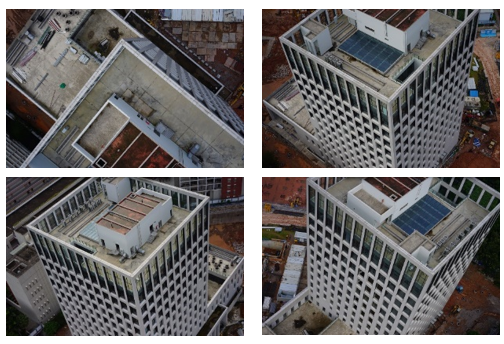}
		\label{fig:figure14-a}
	}
	\subfloat[optimized views photogrammetry]{
		\includegraphics[width=0.4\textwidth]{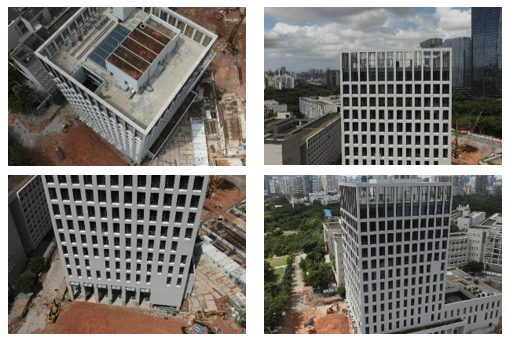}
		\label{fig:figure14-b}
	}
	\caption{The illustration of recorded images from oblique and optimized views photogrammetry.}
	\label{fig:figure14}
\end{figure*}

For the analysis of trajectory planning, Figure \ref{fig:figure13} shows the UAV paths by using oblique photogrammetry and optimized views photogrammetry. We can see that oblique photogrammetry generates a fixed-height and regular trajectory, which pays more attention to the image overlap. On the contrary, optimized views photogrammetry focuses more on the geometric characteristics of ground objects and generates the trajectory that surrounds the targets, as shown in Figure \ref{fig:figure13-b}. In other words, optimized views photogrammetry can obtain enough observations of ground objects, especially for fine-scale and hard-to-see structures in complex urban scenes. This can be verified by the collected UAV images from oblique and optimized views photogrammetry, as presented in Figure \ref{fig:figure14}.

To further analyze the different principles for trajectory planning between oblique and optimized views photogrammetry, we analyze the metric of model point reconstructability. From the first test site, some evenly distributed model points are first selected, which locates on both building tops and facades. Under corresponding view points, the reconstructability of these selected model points can be computed based on Equation \ref{eq:2}, as presented in Section \ref{sec:2}. The calculated reconstructability of model points is quantified into six levels with their values decreasing from high level to low level. Noticeably, level six indicates that the model points can not be observed by any view points. Table \ref{tab:table2} shows the statistic of model point reconstructability of the first test site. It is shown that optimized views photogrammetry has 82.46\% model points with the level one reconstructability, which means that 82.46\% of model points can be well observed. On the contrary, for oblique photogrammetry, the proportion of highest reconstructability is 51.61\%. The main reason is that many model points obtain the reconstructability in levels four and five, whose proportions are 11.79\% and 23.69\%, respectively. In other words, there is a large proportion of model points that can not obtain enough observations under the collected images from the generated trajectory.

\begin{table*}[!ht]
	\centering
	\caption{The statistic of model point reconstructability of the first test site.}
	\label{tab:table2}
	\makebox[\linewidth]{
		\begin{tabular}{llllllll}
			\toprule
			\multirow{2}{*}{Method} & \multirow{2}{*}{Metric} & \multicolumn{6}{l}{Reconstructability level} \\
			\cline{3-8}
			&  & \colorbox{blue}{I} & \colorbox{cyan}{II} & \colorbox{green}{III} & \colorbox{yellow}{IV} & \colorbox{red}{V} & VI \\
			\midrule
			\multirow{2}{*}{Oblique} & \#views & 464 & 26 & 34 & 106 & 213 & 56 \\
			& Percent & 51.61\% & 2.89\% & 3.78\% & 11.79\% & 23.69\% & 6.23\% \\
			\multirow{2}{*}{Optimized} & \#views & 898 & 3 & 3 & 33 & 106 & 46 \\
			& Percent & 82.46\% & 0.28\% & 0.28\% & 3.03\% & 9.73\% & 4.22\% \\
			\bottomrule
		\end{tabular}
	}
\end{table*}

\begin{figure*}[t!]
	\centering
	\subfloat[oblique photogrammetry]{
		\includegraphics[width=0.4\textwidth]{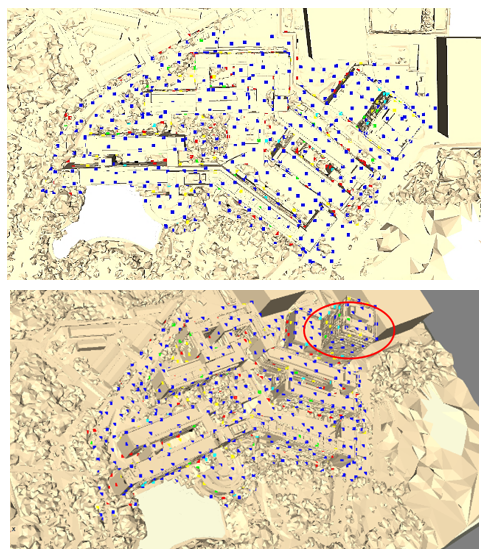}
		\label{fig:figure15-a}
	}
	\subfloat[optimized views photogrammetry]{
		\includegraphics[width=0.4\textwidth]{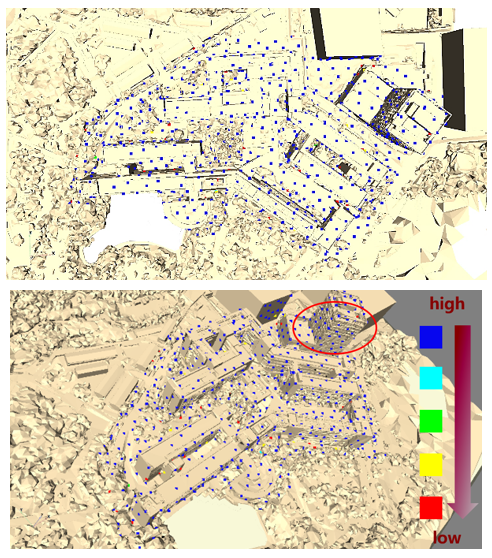}
		\label{fig:figure15-b}
	}
	\caption{The visual comparison of model point reconstructability of the first test site.}
	\label{fig:figure15}
\end{figure*}

For a visual comparison, Figure \ref{fig:figure15} shows the distribution of the selected model points and renders their reconstructability by using different colors. We can observe that for oblique photogrammetry, the model points on the plain plane, e.g., ground or building top, have high reconstructability values; other model points, especially for those on building façade, have very small reconstructability values, as shown in Figure \ref{fig:figure15-a}. On the contrary, for optimized views photogrammetry, the model points with high reconstructability values are evenly distributed over the whole test site, including both building tops and facades, as shown in Figure \ref{fig:figure15-b}. This indicates that under the view points generated from optimized views photogrammetry, almost all model points are well observed and would be well reconstructed in the subsequent 3D modeling.

\subsection{Precision analysis for image orientation and 3D reconstruction}
\label{sec:4.3}

To evaluate the precision of optimized views photogrammetry, both relative BA without GCPs and an absolute BA with GCPs are conducted in this section. Besides, 3D modeling has also been used to evaluate the quality of reconstructed mesh models. The precision test is executed by using datasets collected from oblique and optimized views photogrammetry. In this section, the BA experiments are conducted by using the Bentley ContextCapture software with the version number V4.4.10. All tests are conducted on an Intel Core i7-8700 PC on the Windows platform with 32 GB memory, a 3.19 GHz CPU, and a 6 GB NVIDIA GeForce GTX 1060 graphics card. Due to the high computation costs and large memory consumptions, 3D modeling tests are executed on a high-performance computing cluster.

\subsubsection{Relative BA without GCPs}
\label{sec:4.3.1}

Relative BA without GCPs is first conducted to evaluate the relative accuracy of image orientation. In this test, four metrics are utilized for performance evaluation, which include efficiency, tie-point number, completeness, and precision. The metric efficiency is quantified by the time costs consumed in image matching and orientation; the metric tie-point number is computed by using the median number and the total number of 3D points that are resumed in BA optimization, respectively; the metrics completeness and precision respectively indicate the number of connected images and the re-projection error after BA optimization. Table \ref{tab:table3} lists the statistical results of relative BA by using the four metrics. We can see that: (1) the time costs of optimized views photogrammetry are higher than that of oblique photogrammetry because the number of images captured by optimized views photogrammetry is larger than that of oblique photogrammetry, which are 3620 and 4030 for these two methods, respectively; (2) although more images have been captured in optimized views photogrammetry, the number of tie-points is less than that of oblique photogrammetry. The main reason is that the imaging direction camera is adjusted in optimized views photogrammetry according to the normal vectors of ground objects, which causes more low-texture regions in the collected images, as illustrated in Figure \ref{fig:figure14-b}; (3) the metrics of completeness and precision are better in optimized views photogrammetry, which connects all collected images.

\begin{table}[!ht]
	\centering
	\caption{The statistic of relative BA without GCPs for the first test site.}
	\label{tab:table3}
	\makebox[\linewidth]{
		\begin{tabular}{llllll}
			\toprule
			\multirow{2}{*}{Method} & \multirow{2}{*}{\begin{tabular}[c]{@{}l@{}}Efficiency \\ (Minutes)\end{tabular}} & \multicolumn{2}{l}{Tie-point number} & \multirow{2}{*}{Completeness} & \multirow{2}{*}{\begin{tabular}[c]{@{}l@{}}Precision\\  (pixels)\end{tabular}} \\
			\cline{3-4}
			&  & Median & Total &  &  \\
			\midrule
			Oblique & 50.8 & 948 & 654,468 & 3615/3620 & 0.69 \\
			Optimized & 60.1 & 986 & 558,751 & 4030/4030 & 0.62 \\
			\bottomrule
		\end{tabular}
	}
\end{table}

\begin{figure}[t!]
	\centering
	\subfloat[oblique photogrammetry]{
		\includegraphics[width=0.23\textwidth]{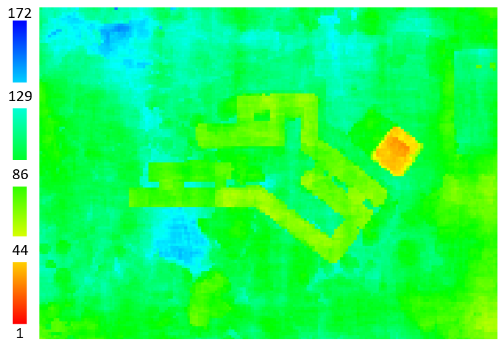}
		\label{fig:figure16-a}
	}
	\subfloat[optimized views photogrammetry]{
		\includegraphics[width=0.23\textwidth]{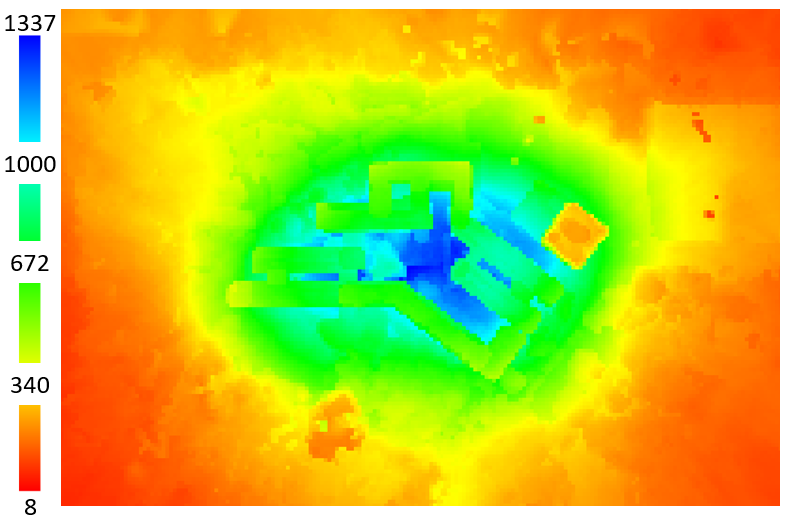}
		\label{fig:figure16-b}
	}
	\caption{The comparison of image ground coverage of the first test site.}
	\label{fig:figure16}
\end{figure}

\begin{figure}[t!]
	\centering
	\subfloat[oblique photogrammetry]{
		\includegraphics[width=0.23\textwidth]{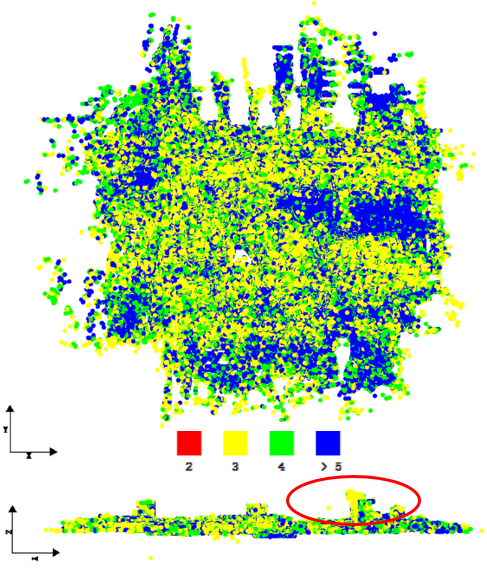}
		\label{fig:figure17-a}
	}
	\subfloat[optimized views photogrammetry]{
		\includegraphics[width=0.23\textwidth]{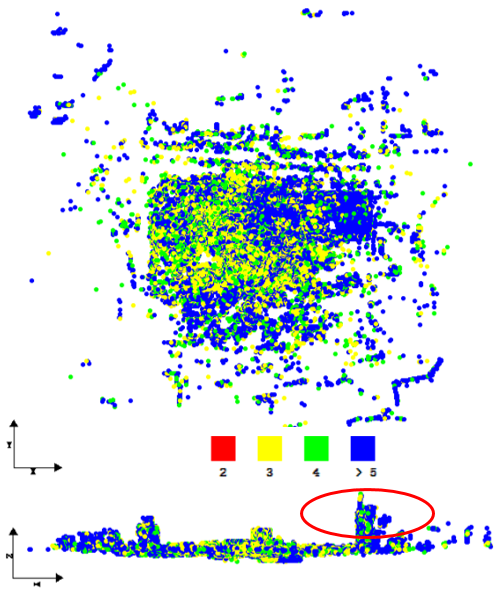}
		\label{fig:figure17-b}
	}
	\caption{The comparison of tie-point length distribution of the first test site.}
	\label{fig:figure17}
\end{figure}

To further evaluate the performance of optimized views photogrammetry, image ground coverage and tie-point length distribution are also analyzed and rendered in this test. Image ground coverage indicates the number of images that can see a specified ground point; tie-point length distribution indicates the number of images that can observe the tie-point. Figure \ref{fig:figure16} and Figure \ref{fig:figure17} show the image ground coverage and tie-point length distribution by using the relative BA results. It clearly shows that oblique photogrammetry uses very evenly distributed coverage for both interesting and non-interesting regions, and it does not consider the geometric structure of ground objects, as shown in Figure \ref{fig:figure16-a}, because regular trajectory path and fixed imaging direction have been used in the data acquisition; on the contrary, UAV trajectory has been optimized to adapt the characteristics of ground objects, which can be observed from the buildings-occupied blue region and the non-interesting red region. Due to the advantages of optimized views photogrammetry, many more and long tie-points can be generated on building facades, which can be verified by the results in Figure \ref{fig:figure17}.

\subsubsection{Absolute BA with GCPs}
\label{sec:4.3.2}

\begin{figure}[ht!]
	\centering
	\subfloat[horizontal direction]{
		\includegraphics[width=0.4\textwidth]{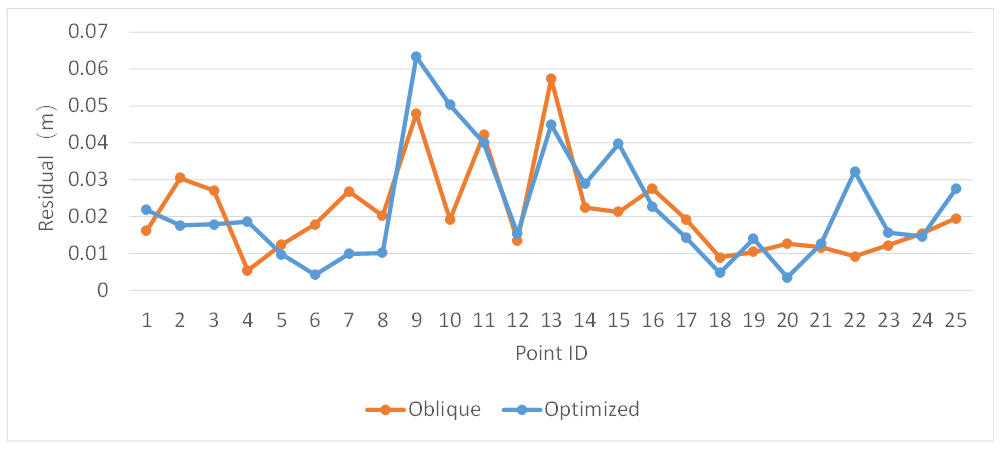}
		\label{fig:figure18-a}
	} \\
	\subfloat[vertical direction]{
		\includegraphics[width=0.4\textwidth]{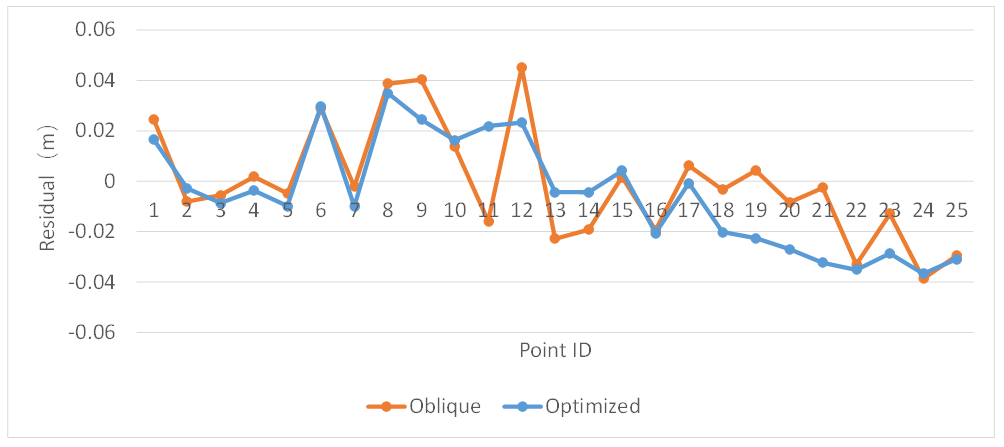}
		\label{fig:figure18-b}
	}
	\caption{The residual distribution of the absolute BA test.}
	\label{fig:figure18}
\end{figure}

Absolute BA with GCPs is conducted to evaluate the geo-referencing accuracy of image orientation. In this test, three GCPs, labeled K02, K04, and K10 as shown in Figure \ref{fig:figure10}, are used as control points in BA optimization, and the other 25 GCPs are used as checkpoints (CPs) for accuracy assessment. Table \ref{tab:table4} shows the residual statistics of the absolute BA test. In this test, the residual is calculated as the coordinate difference between measured model points and checkpoints. We can see that: (1) the maximum residuals in horizontal and vertical directions are 0.063 m and 0.035 m, respectively, for optimized views photogrammetry; (2) although oblique photogrammetry has smaller residual in the horizontal direction, the residual reaches 0.45 m in the vertical direction; (3) considering RMSE (root mean square error), comparative BA accuracy has been achieved between these two methods, which can also be demonstrated by the consistent residual distribution as presented in Figure \ref{fig:figure18}. In a conclusion, although more attention has been paid to the ground objects instead of the image connection network in oblique photogrammetry, reliable image connection can also be established in optimized views photogrammetry to ensure the geo-referencing precision.

\begin{table}[!t]
	\centering
	\caption{Residual statistics for absolute BA test (unit in meters).}
	\label{tab:table4}
	\makebox[\linewidth]{
		\begin{tabular}{lllllll}
			\toprule
			\multirow{2}{*}{Method} & \multicolumn{2}{l}{Min} & \multicolumn{2}{l}{Max} & \multicolumn{2}{l}{RMSE} \\
			\cline{2-7}
			& XY & Z & XY & Z & XY & Z \\
			\midrule
			Oblique & 0.005 & -0.039 & 0.057 & 0.045 & 0.024 & 0.022 \\
			Optimized & 0.004 & -0.037 & 0.063 & 0.035 & 0.026 & 0.022 \\
			\bottomrule
		\end{tabular}
	}
\end{table}

\subsubsection{Quality of 3D mesh models}
\label{sec:4.3.3}

3D mesh models can be created based on MVS dense matching and point cloud meshing. In this section, TLS point clouds in the first test site are used as ground-truth data to evaluate the quality of generated 3D mesh models. For an overall comparison, discrepancies between mesh models and point clouds are first calculated, and two metrics are used for performance evaluation, including precision and completeness. The metric precision indicates the error of mesh model points to ground-truth point clouds. It is calculated by sorting discrepancies in ascending order and finding three thresholds such that there are 50\%, 70\%, and 90\% model points, respectively, whose discrepancies are below the corresponding threshold. The metric completeness indicates the percentage of mesh model points whose discrepancies are less than the given thresholds, i.e., 0.01 m, 0.1 m, 0.5 m, and 1.0 m. The statistical results are listed in Table \ref{tab:table5}, and the results show that the precision of optimized views photogrammetry is higher than oblique photogrammetry, whose values are 0.041 m, 0.077 m, and 0.159 m under the three thresholds, respectively. Considering completeness, when a threshold is set as 0.01 m, the performance is not satisfied because the reconstruction cannot reach this accuracy level. By using other thresholds, the average completeness of optimized views photogrammetry is approximately 14.7\% higher than that of oblique photogrammetry.

\begin{table*}[!t]
	\centering
	\caption{Statistical results of precision and completeness of Mesh models.}
	\label{tab:table5}
	\makebox[\linewidth]{
		\begin{tabular}{llllllll}
			\toprule
			\multirow{2}{*}{Method} & \multicolumn{3}{l}{Precision (m)} & \multicolumn{4}{l}{Completeness} \\
			\cline{2-8}
			& 50\% & 70\% & 90\% & 0.01 m & 0.1 m & 0.5 m & 1.0 m \\
			\midrule
			Oblique & 0.048 & 0.086 & 0.172 & 0.15\% & 30.36\% & 60.07\% & 72.74\% \\
			Optimized & 0.041 & 0.077 & 0.159 & 0.27\% & 43.98\% & 77.06\% & 86.16\% \\
			\bottomrule
		\end{tabular}
	}
\end{table*}

\begin{table*}[!t]
	\centering
	\caption{Statistical results for the residual of building facades.}
	\label{tab:table6}
	\makebox[\linewidth]{
		\begin{tabular}{lllllll}
			\toprule
			\multirow{2}{*}{Facade} & \multicolumn{2}{l}{Max (m)} & \multicolumn{2}{l}{Mean (m)} & \multicolumn{2}{l}{Std.dev. (m)} \\
			\cline{2-7}
			& Oblique & Optimized & Oblique & Optimized & Oblique & Optimized \\
			\midrule
			1 & 1.12 & 1.10 & 0.09 & 0.03 & 0.19 & 0.12 \\
			2 & 2.70 & 2.42 & 0.17 & 0.11 & 0.33 & 0.25 \\
			3 & 2.26 & 2.28 & 0.16 & 0.03 & 0.27 & 0.12 \\
			\bottomrule
		\end{tabular}
	}
\end{table*}

\begin{figure}[!ht]
	\centering
	\includegraphics[width=0.45\textwidth]{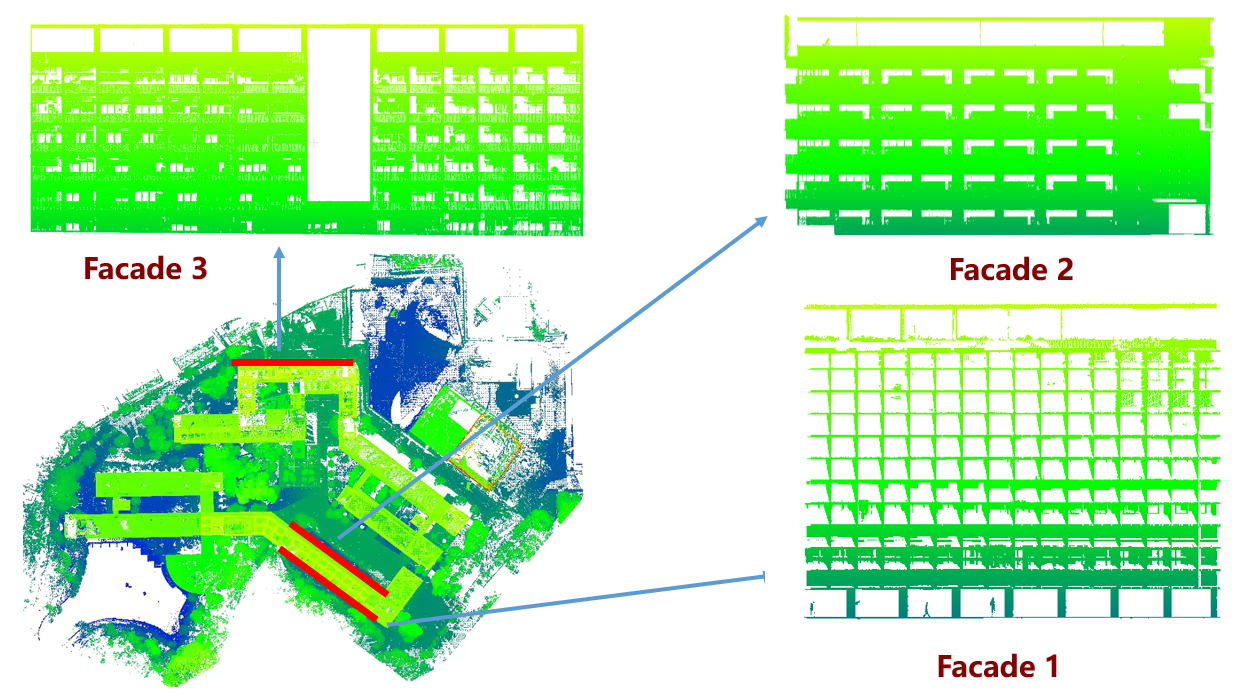}
	\caption{The distribution of building facades and corresponding point clouds.}
	\label{fig:figure19}
\end{figure}

\begin{figure*}[ht!]
	\centering
	\subfloat[facade 1 – oblique]{
		\includegraphics[width=0.4\textwidth]{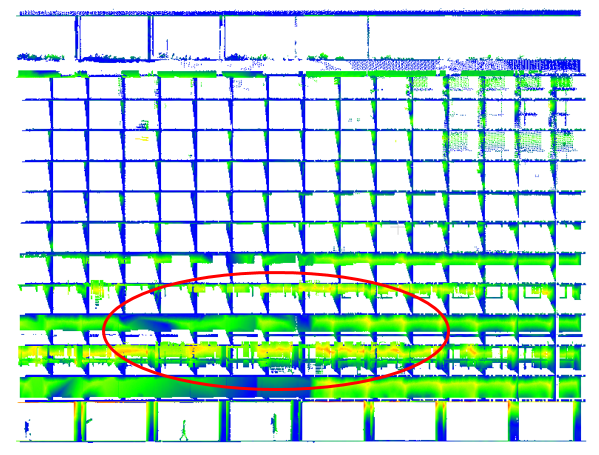}
		\label{fig:figure20-a}
	}
	\subfloat[facade 1 – optimized]{
		\includegraphics[width=0.4\textwidth]{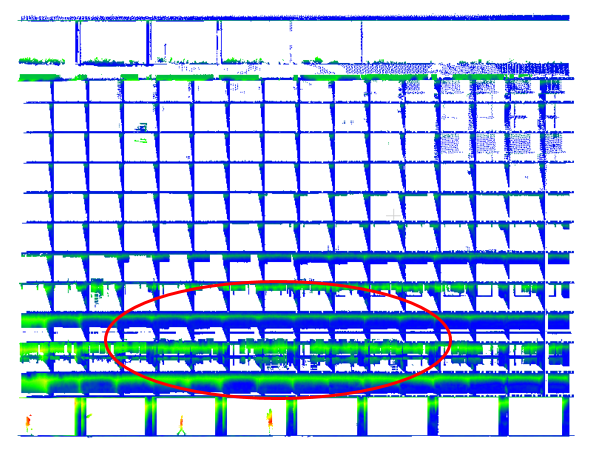}
		\label{fig:figure20-b}
	} \\
	\subfloat[facade 2 – oblique]{
		\includegraphics[width=0.4\textwidth]{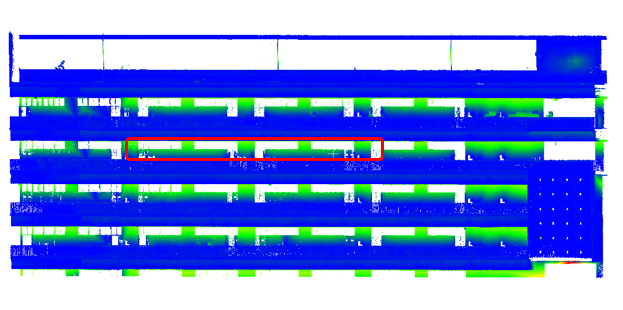}
		\label{fig:figure20-c}
	}
	\subfloat[facade 2 – optimized]{
		\includegraphics[width=0.4\textwidth]{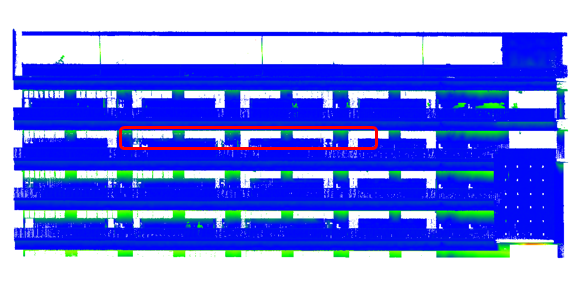}
		\label{fig:figure20-d}
	} \\
	\subfloat[facade 3 – oblique]{
		\includegraphics[width=0.4\textwidth]{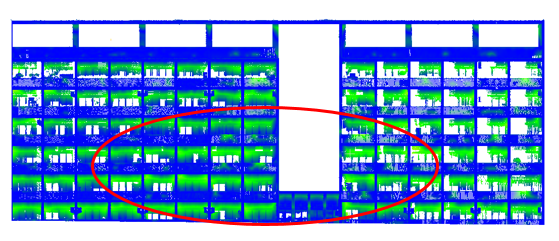}
		\label{fig:figure20-e}
	}
	\subfloat[facade 3 – optimized]{
		\includegraphics[width=0.4\textwidth]{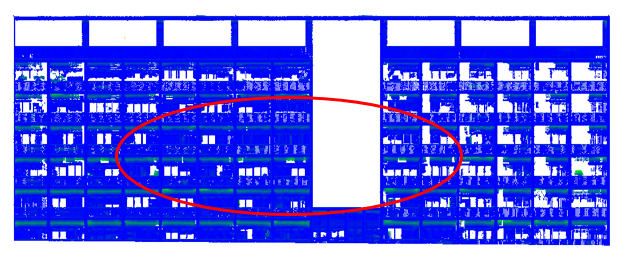}
		\label{fig:figure20-f}
	}
	\caption{The distribution of errors of building facades.}
	\label{fig:figure20}
\end{figure*}

For further comparison, three building facades are selected and trimmed from mesh models, and a comparison with TLS point clouds is then conducted. Figure \ref{fig:figure19} shows the distribution of the selected 3 building facades and the corresponding point clouds. Facade 1 contains a large number of glass windows, and the laser point clouds are mainly distributed on the outer wall of the building; facade 2 contains concave corridors along with balconies, and the laser point clouds are evenly distributed in this region; facade 3 contains a large number of glass windows and concave balconies. By using the trimmed point clouds and reconstructed mesh models, the residual is calculated as the distance between these laser point clouds and reconstructed mesh models. The statistical results are listed in Table \ref{tab:table6}. Besides, Figure \ref{fig:figure20} shows the error distribution map of the building facades, and the error statistical histogram is attached in the upper-left corner.

The results show that except for facade 3, the reconstruction model of optimized views photogrammetry is better than that of oblique photogrammetry. Especially, for facades 1 and 3 with serious occlusions, optimized views photogrammetry can significantly reduce the reconstruction error, which is about 3 to 5 times higher than that of oblique photogrammetry. The main reason is that facades 1 and 3 contain many concave windows, which leads to many hard-to-see regions in oblique photogrammetry, as shown in Figure \ref{fig:figure20-a} and Figure \ref{fig:figure20-e}. Compared with oblique photogrammetry, optimized views photogrammetry significantly improves the accuracy of reconstructed models and reduces the problem of insufficient image acquisition in the hard-to-see areas of building facades, as demonstrated by the error distribution in Figure \ref{fig:figure20-b}, Figure \ref{fig:figure20-d} and Figure \ref{fig:figure20-f}.

\subsection{Engineer application for a large-scale site in Qingdao}
\label{sec:4.4}

To verify the capability in engineering application, a large-scale site has been selected, which locates in Qingdao city, China. The overall coverage of this site is shown in Figure \ref{fig:figure12}. Based on the workflow of optimized views photogrammetry, the rough model of the site is first generated by using the oblique UAV dataset. A trajectory path is then created for each subsite as described in Section \ref{sec:4.1}. To achieve efficient acquisition, a multi-UAV cooperation model has been utilized on this site. After UAV image collection, 3D models of the whole test site are reconstructed. The details of the engineering application are presented as follows.

\subsubsection{Rough model generation and trajectory planning}
\label{sec:4.4.1}

\begin{figure*}[!ht]
	\centering
	\includegraphics[width=0.8\textwidth]{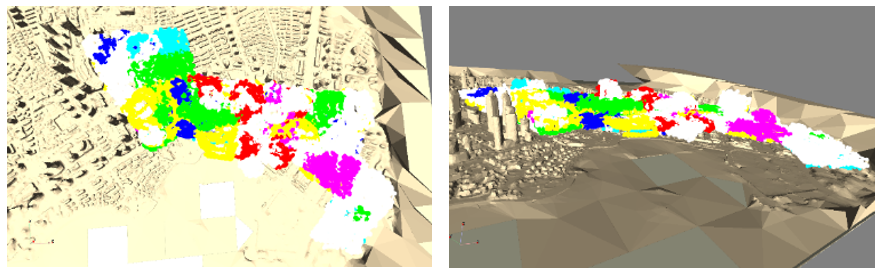}
	\caption{The overall trajectory generated based on the rough model.}
	\label{fig:figure21}
\end{figure*}

Rough models are the basis for trajectory planning in optimized views photogrammetry. In this site, oblique photogrammetry is utilized for UAV image acquisition, and there are a total number of 2433 images recorded by using a classical Penta-view imaging system. The rough model is then created by RealityCapture software, and the computer configuration is a 3.7 GHz Intel Core i9-10900X processor, and a 12 G RTX3080 graphics card. 40 hours have been consumed for creating the rough model, which is then used in optimized views photogrammetry to plan UAV trajectories for the fine-scale data acquisitions. According to the division strategy, as presented in Section \ref{sec:4.1}, the generated UAV trajectories of all zones are illustrated in Figure \ref{fig:figure21}, in which the trajectories are rendered by varying colors.

\begin{figure*}[ht!]
	\centering
	\subfloat[zone 2 for high buildings]{
		\includegraphics[width=0.8\textwidth]{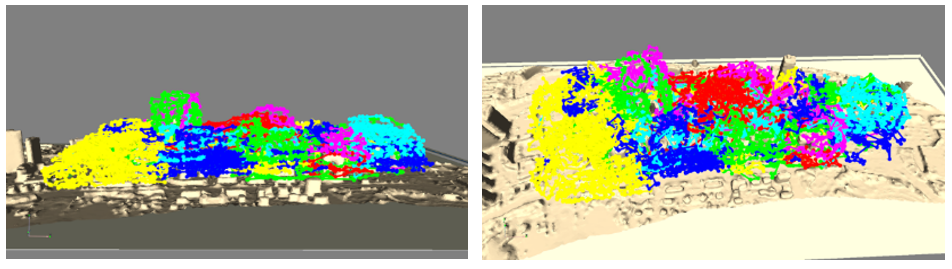}
		\label{fig:figure22-a}
	} \\
	\subfloat[zone 3 with extremely large height difference]{
		\includegraphics[width=0.8\textwidth]{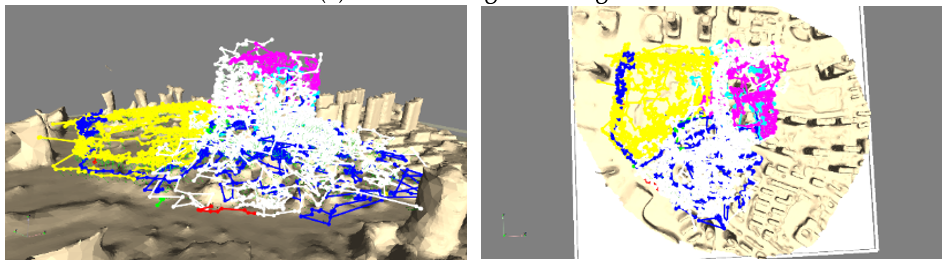}
		\label{fig:figure22-b}
	} \\
	\subfloat[zone 4 for low-height buildings]{
		\includegraphics[width=0.8\textwidth]{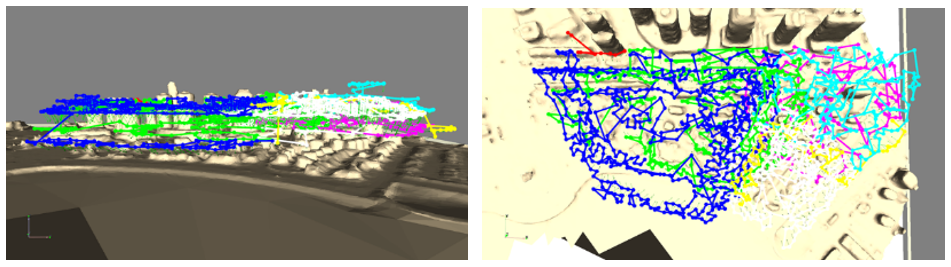}
		\label{fig:figure22-c}
	}
	\caption{The illustration of trajectory paths for zones with varying characteristics.}
	\label{fig:figure22}
\end{figure*}

For further analysis, Figure \ref{fig:figure22} shows the trajectories of zones 2, 3, and 4. Zone 2 is a central business distinct with many high buildings, and the safe-flying zone between buildings is very limited. The trajectories are generated around some high buildings. In zone 3, a very high building locates at the center of this region, which is surrounded by some low buildings. For trajectory planning, conventional oblique photogrammetry would just consider the height of the highest building. Zone 4 is covered by a low-resident building. These three zones stand for the classical ground coverage for data acquisition in urban scenes. In these zones, suitable trajectories can be generated based on the optimized views photogrammetry.

\subsubsection{Multi-UAV cooperation for data acquisition}
\label{sec:4.4.2}

Multi-UAV cooperation is optimal for data acquisition in the large-scale site due to two main reasons. On the one hand, multi-rotor UAVs are widely used for data acquisition in complex urban scenes because of their operation safety and acquisition flexibility. However, the endurance of almost all market-available UAVs is very limited, such as less than 25 min for the DJI Phantom 4 RTK; on the other hand, by using optimized views photogrammetry, the trajectories are relatively longer than that of classical oblique photogrammetry. In trajectory planning, optimized views photogrammetry has also considered the multi-UAV task mode in practice. Thus, the generated trajectories have no spatial overlap region, which can ensure the operation safety of multi-UAV cooperation.

In this test site, six DJI Phantom 4 RTK UAVs have been used for the parallel acquisition. For the six zones, the statistical result in data acquisition is listed in Table \ref{tab:table7}. During data acquisition, the speed of UAVs is set as 5 m/s, and the image record interval is configured as 3 s. In Table \ref{tab:table7}, the number of view-point images and interpolated images indicate the images that are recorded at the planning view points and the intermediate points between two view points, respectively. The number of recorded images is the total number of images in the zone. We can see that the most time costs are consumed in zone 2 due to its complex environment, and a total number of 78640 images are collected.

\begin{table*}[!t]
	\centering
	\caption{Statistical results of optimized views photogrammetry implementation}
	\label{tab:table7}
	\makebox[\linewidth]{
		\begin{tabular}{lllllll}
			\toprule
			\multirow{2}{*}{Metric} & \multicolumn{6}{l}{Zone number} \\
			\cline{2-7}
			& 1 & 2 & 3 & 4 & 5 & 6 \\
			\midrule
			Trajectory length (m) & 94757 & 173631 & 60254 & 23949 & 90576 & 92249 \\
			\# trajectory & 74 & 236 & 76 & 17 & 94 & 60 \\
			Planned time cost (min) & 152 & 243 & 55 & 36 & 106 & 137 \\
			Actual time cost (min) & 316 & 579 & 201 & 80 & 302 & 308 \\
			\# view point image & 9030 & 18575 & 5957 & 1858 & 8112 & 6589 \\
			\# interpolated image & 6320 & 11580 & 4020 & 1600 & 6040 & 6160 \\
			\# recorded image & 14443 & 27394 & 9245 & 2838 & 12445 & 12275 \\
			\bottomrule
		\end{tabular}
	}
\end{table*}

\subsubsection{3D model reconstruction of Qingdao city}
\label{sec:4.4.3}

\begin{figure*}[ht!]
	\centering
	\subfloat[image orientation of the whole site]{
		\includegraphics[height=0.2\textwidth]{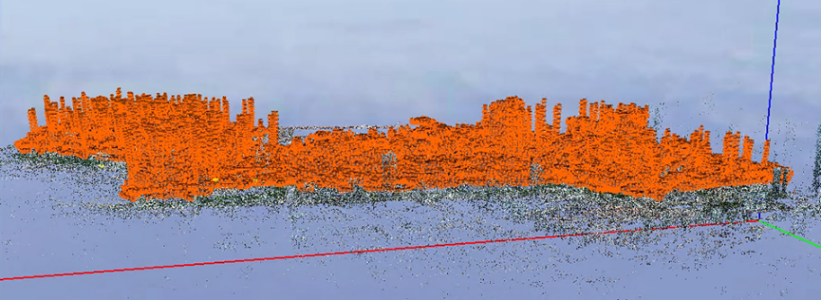}
		\label{fig:figure23-a}
	}
	\subfloat[3D reconstruction of the whole site]{
		\includegraphics[height=0.2\textwidth]{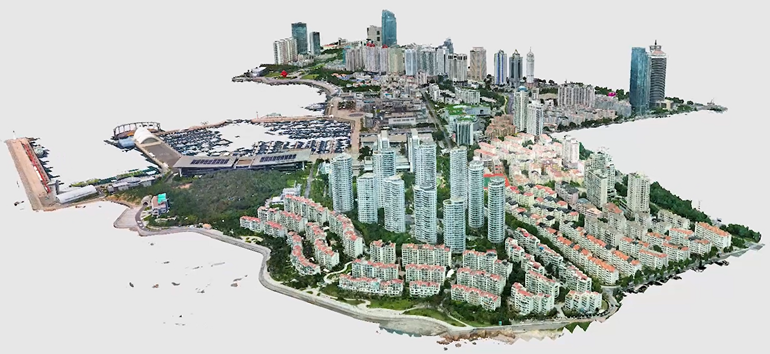}
		\label{fig:figure23-b}
	}
	\caption{The results of image orientation and 3D reconstruction of the whole site.}
	\label{fig:figure23}
\end{figure*}

3D reconstruction is finally conducted to generate models after outdoor data acquisition. The Bentley ContextCapture software (version 20210824) has been adopted in this study. For image orientation, only one computing node is used, which is configured with one Intel(R) Core(TM) i5-9600KF, 128 G memory, and one NVIDIA GTX1660 super graphic card. For 3D modeling, a computing cluster with 25 nodes has been used for efficiency improvement. Based on these configurations, the time costs of image orientation and 3D reconstruction are 46 hours and 156 hours, respectively, for the whole site. Figure \ref{fig:figure23} shows the results of image orientation and 3D reconstruction. We can see that the recorded images can be successfully connected, and the reconstructed model covers the whole test site.

\begin{figure*}[ht!]
	\centering
	\subfloat[3D model and local details of zone 2.]{
		\includegraphics[height=0.3\textwidth]{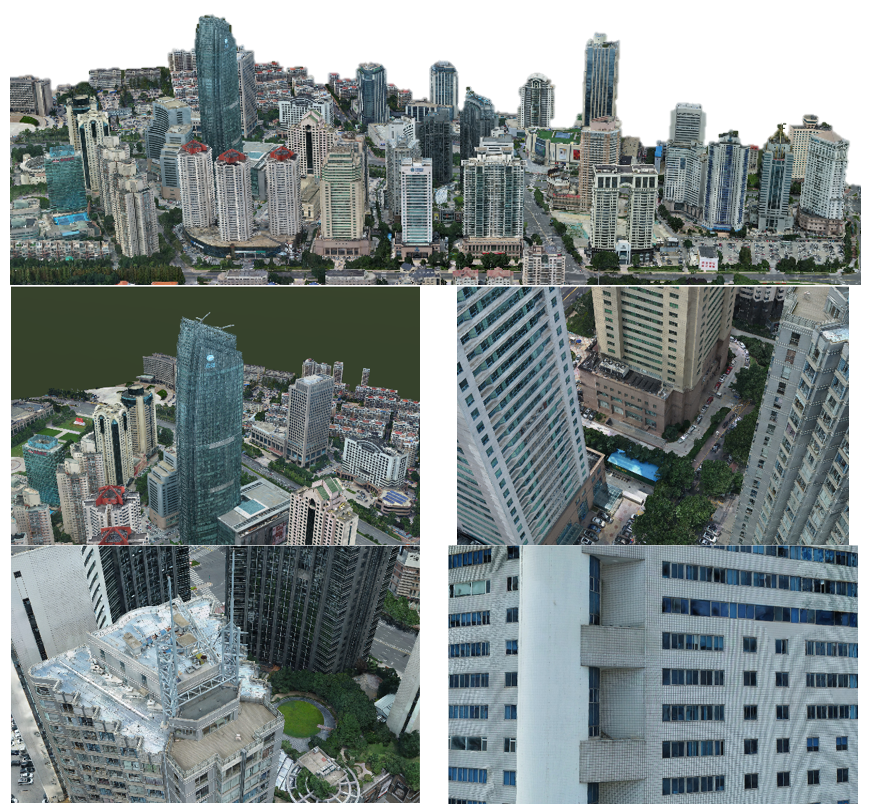}
		\label{fig:figure24-a}
	}
	\subfloat[3D model and local details of zone 3.]{
		\includegraphics[height=0.3\textwidth]{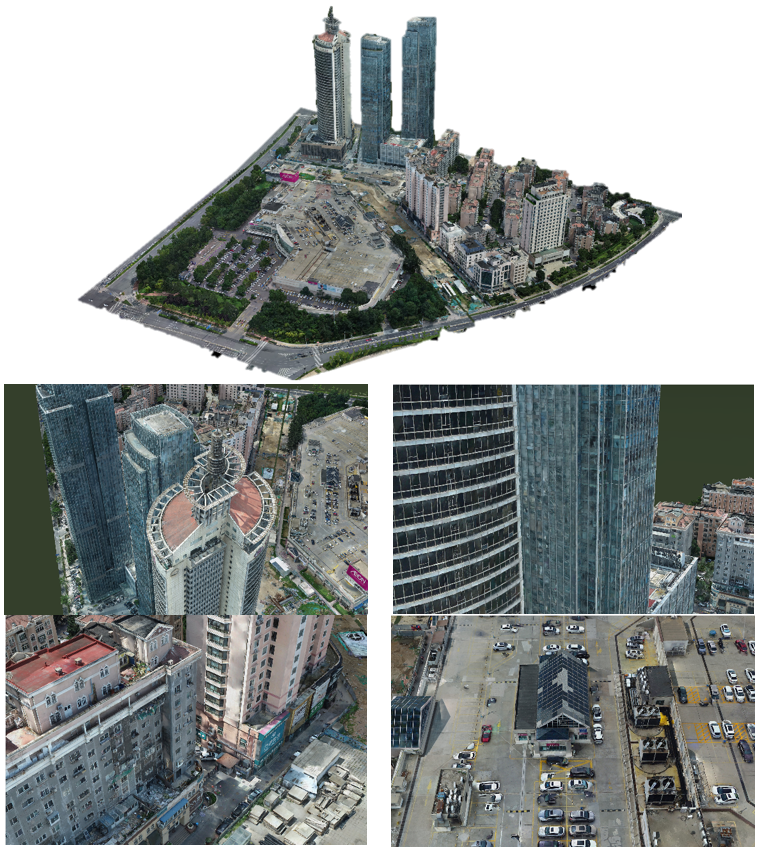}
		\label{fig:figure24-b}
	}
	\subfloat[3D model and local details of zone 4.]{
		\includegraphics[height=0.3\textwidth]{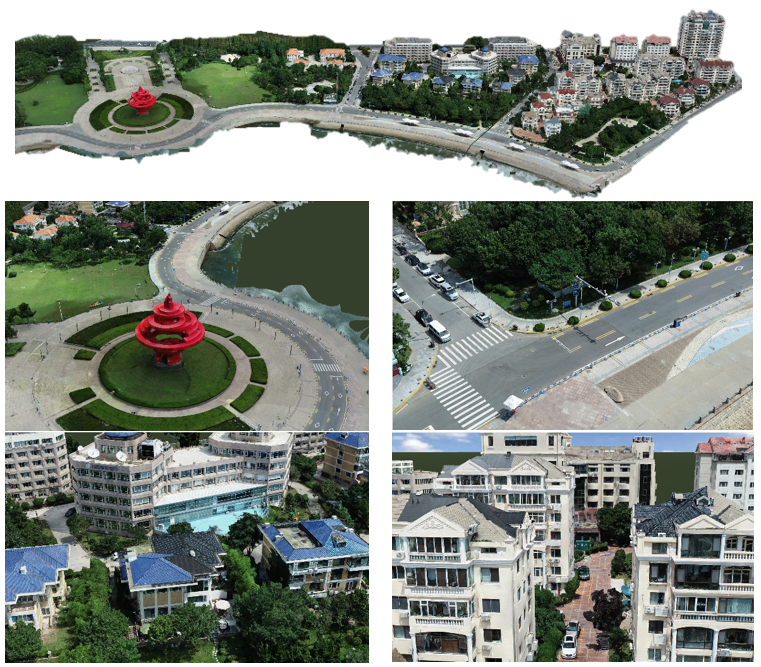}
		\label{fig:figure24-c}
	}
	\caption{3D models and local details of the three zones.}
	\label{fig:figure24}
\end{figure*}

For a further visual interpretation, 3D models of zones 2, 3, and 4 are presented in Figure \ref{fig:figure24}. It is shown that for these three classical urban environments, optimized views photogrammetry can collect enough images for accurate reconstruction. Figure \ref{fig:figure24-a} shows the results of the 3D model in the high-rise building area, which includes glass curtain walls, ground podiums between narrow-spaced buildings, steel-frame structure on the building top, and low building facades. Figure \ref{fig:figure24-b} is the models in the area with large height variance. This site includes special-shaped structures on building tops, glass curtain walls of building facades, low residential buildings, and car parks. Figure \ref{fig:figure24-c} is a residential areas with low-rise buildings, which includes landmarks, traffic roads, villa areas, and multi-level residential buildings. For these three classical zones, optimized views photogrammetry can generate high-quality 3D models. From the reconstruction results, we can conclude that optimized views photogrammetry provides a reliable and powerful solution for 3D modeling in complex urban scenes.

\section{Conclusions}
\label{sec:5}

UAVs have become one of the widely used RS platforms and play a critical role in the construction of smart cities. However, due to the complex environment in urban scenes, safe and efficient data acquisition brings great challenges to 3D modeling and scene updating. How to achieve optimal trajectory planning of UAV platforms and accurate data collection of on-board cameras has become one of the non-trivial problems in the 3D China construction of urban cities. Different from oblique photogrammetry, optimized views photogrammetry uses an urban rough model to generate and optimize the trajectory of UAVs, which combines initial viewpoint generation with dense sampling and viewpoint optimization under the constraints of model point reconstructability and view point redundancy.

According to the principle of optimized views photogrammetry, this study conducts a precision analysis of 3D models by using UAV images of optimized views photogrammetry and executes a large-scale study in the urban region of Qingdao city, China, to demonstrate the performance of optimized views photogrammetry in engineering applications. By using both GCPs for image orientation precision analysis and TLS point clouds for model quality analysis, experimental results demonstrate that optimized views photogrammetry could construct stable image connection networks and achieves comparative relative and absolute image orientation accuracy. Due to its accurate acquisition data strategy, the quality of the 3D mesh model is significantly improved. Especially for seriously occluded areas, optimized views photogrammetry can achieve 3 to 5 times of accuracy improvement. Most importantly, the case study for the large-scale 3D urban modeling in Qingdao city verifies that optimized views photogrammetry can provide a valuable and powerful solution for 3D reconstruction in complex urban scenes. Thus, future studies would attempt to promote the application of optimized views photogrammetry in other related fields, such complex structure inspection \cite{khaloo2018unmanned,shang2020co}, cultural heritage documentation \cite{bakirman2020implementation,pan2019three}.

\section*{Acknowledgment}
\label{Acknowledgment}

The authors would like to thank the anonymous reviewers and editors, whose comments and advice improve the quality of the work. This research was funded by the project of Qingdao real scene 3D construction (ZFCG2021000043), Basic geographic information data construction project of Qingdao high tech Zone (GXCG2020000113), Natural Science Foundation of Guangdong Province (2020A0505100064), Shenzhen Key Project of Science and Technology Innovation (JCYJ20210324120213036), and the National Natural Science Foundation of China (42001413).



\ifCLASSOPTIONcaptionsoff
  \newpage
\fi



%
%
%

\bibliographystyle{IEEEtran}
\nocite{*}
\bibliography{mybibfile}

%

\begin{IEEEbiography}[{\includegraphics[width=1in,height=1.25in,clip,keepaspectratio]{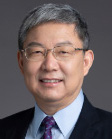}}]{Qingquan Li}
received the Ph.D. degree in Geographic
Information System (GIS) and photogrammetry
from the Wuhan Technical University of Surveying and Mapping, Wuhan, China, in 1998.
From 1988 to 1996, he was an Assistant Professor with Wuhan University, Wuhan, where he became an Associate Professor, in 1996, and has been a Professor, since 1998. He is currently a Professor of Shenzhen University, Shenzhen, China; a Professor with the State Key Laboratory of Information Engineering in Surveying, Mapping and Remote Sensing, Wuhan University; and the Director of Shenzhen Key Laboratory of Spatial Smart Sensing and Service, Shenzhen. His research interests include intelligent transportation systems, 3-D and dynamic data modeling, and pattern
recognition.

Dr. Li is an Academician of International Academy of Sciences for Europe and Asia (IASEA), an Expert in Modern Traffic with the National 863 Plan, and an Editorial Board Member of the Surveying and Mapping Journal and the Wuhan University Journal—Information Science Edition.
\end{IEEEbiography}

\begin{IEEEbiography}[{\includegraphics[width=1in,height=1.25in,clip,keepaspectratio]{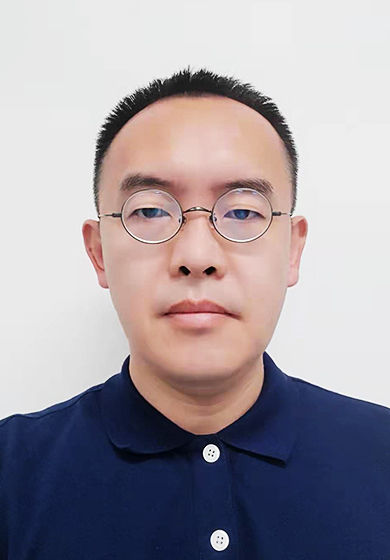}}]{Wenshuai Yu}
received the bachelor's degree in aerospace photogrammetry and the master's degree in photogrammetry and remote sensing from the Institute of Survey and Mapping, Zhengzhou, China, in 2003 and 2006, respectively. He received the Ph.D. degree in photogrammetry and remote sensing from the Institute of Survey and Mapping, Zhengzhou, China.

He is currently an associate researcher with the College of Civil and Transportation Engineering, Shenzhen University, and a principal investigator with the Guangdong Laboratory of Artificial Intelligence and Digital Economy, Shenzhen, China. His reseach interests include UAV photogrammetry, intelligent 3D spatial perception, SLAM and 3D Reconstruction.
\end{IEEEbiography}

\begin{IEEEbiography}[{\includegraphics[width=1in,height=1.25in,clip,keepaspectratio]{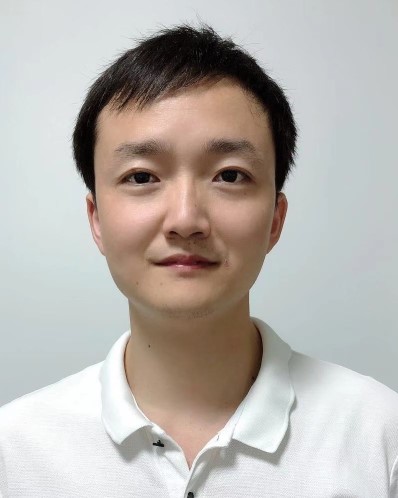}}]{San Jiang}
received the B.S. degree in remote sensing science and technology from Wuhan University in 2010, and the M.Sc. and Ph.D. degrees in photogrammetry and remote sensing from Wuhan Univeristy in 2012 and 2018, respectively. From 2012 to 2014, he worked as an assistant engineer in Tianjin Institute of Surveying and Mapping. From 2014 to 2015, he joined the LIESMARS (State Key Laboratory of Information Engineering in Surveying, Mapping and Remote Sensing of Wuhan Univeristy) as a research assistant.
	
Currently, he is an associate professor in the School of Computer Science at China University of GeoSciences (Wuhan). His research interests include image matching, SfM-based aerial triangulation, and 3D reconstruction.
\end{IEEEbiography}





\end{document}